\documentclass[accepted,specialissue]{melba}

\usepackage{mwe} 

\usepackage{amsmath,amsfonts}
\usepackage{booktabs}
\usepackage{multirow}
\usepackage[acronym]{glossaries}
\glsdisablehyper
\usepackage[table,xcdraw]{xcolor}
\makeglossaries
\usepackage{float}
\usepackage{xcolor}
\usepackage{xurl}
\usepackage{bookmark}

\newacronym{uq}{UQ}{Uncertainty Quantification}
\newacronym{who}{WHO}{World Health Organization}
\newacronym{mcd}{MCD}{Monte Carlo Dropout}
\newacronym{mcde}{MCDE}{Monte Carlo Deep Ensembles}
\newacronym{idh}{IDH}{Isocitrate Dehydrogenase}
\newacronym{mri}{MRI}{Magnetic resonance imaging}
\newacronym{ai}{AI}{Artificial Intelligence}
\newacronym{dl}{DL}{Deep Learning}
\newacronym{mi}{MI}{Mutual Information}
\newacronym{ece}{ECE}{Expected Calibration Error}
\newacronym{nll}{NLL}{Negative Log-Likelihood}
\newacronym{dsc}{DSC}{Dice Similarity Coefficient}
\newacronym{auc}{AUC}{Receiver Operating Characteristic Area Under the Curve}
\newacronym{u-auc}{U-AUC}{Uncertainty-based Receiver Operating Characteristic Area Under the Curve}
\newacronym{t1w}{T1w}{T1-weighted}
\newacronym{t1wce}{T1wCE}{Contrast-Enhanced T1-weighted}
\newacronym{t2w}{T2w}{T2-weighted}
\newacronym{flair}{FLAIR}{Fluid-Attenuated Inversion Recovery}
\newacronym{cnn}{CNN}{Convolutional Neural Network}
\newacronym{cnns}{CNNs}{Convolutional Neural Networks}
\newacronym{bnns}{BNNs}{Bayesian Neural Networks}
\newacronym{mc}{MC}{Monte Carlo}
\newacronym{de}{DE}{Deep Ensembles}
\newacronym{nn}{NN}{Neural Network}
\newacronym{nns}{NNs}{Neural Networks}
\newacronym{ap}{AP}{Average Precision}
\newacronym{tta}{TTA}{Test-time Augmentation}
\newacronym{el}{EL}{Evidential Learning}
\newacronym{gm}{GM}{Generative Modelling}
\newacronym{cp}{CP}{Conformal Prediction}
\newacronym{aurc}{AURC}{Area Under the Risk--Coverage Curve}

\melbaid{2026:033}  
\doi{10.59275/j.melba.2026-456d}
\melbaauthors{Mosquera Rojas et al}  
\email{g.mosquerarojas@erasmusmc.nl}
\volume{2026}
\firstpageno{674}  
\melbayear{2026}  
\datesubmitted{2026-03-09}  
\datepublished{2026-09-28}  

\melbaspecialissue{Uncertainty for Safe Utilization of Machine Learning in Medical Imaging (UNSURE) 2025}
\melbaspecialissueeditors{Mobarak Hoque, Raghav Mehta, Cheng Ouyang, Chen Qin, Marianne Rakic, Sandy Wells}

\ShortHeadings{Towards Trustworthy AI for Glioma Diagnosis: A Task-Aware Evaluation of Uncertainty Quantification}{Mosquera Rojas et al.}

\title{Towards Trustworthy AI for Glioma Diagnosis: A Task-Aware Evaluation of Uncertainty Quantification}

\author{
	\firstname Gonzalo Esteban \surname Mosquera Rojas\aff{1} \orcid{0009-0008-9901-4141},
	\firstname Sebastian R. \surname van der Voort\aff{2} \orcid{0000-0002-6526-8126}, \firstname Carolin M. \surname Pirkl\aff{3}, \firstname Sandeep \surname Kaushik\aff{4} \orcid{0000-0003-0654-0799}, \firstname Marion \surname Smits\aff{1,5,6} \orcid{0000-0001-5563-2871}, \firstname Stefan \surname Klein\aff{1} \orcid{0000-0003-4449-6784}
}
\affiliations{
	\num 1 \addr Department of Radiology and Nuclear Medicine, Erasmus MC, University Medical Center Rotterdam, Rotterdam, the Netherlands \\
	\num 2 \addr Department of Medical Informatics, Amsterdam UMC, University of Amsterdam,
Amsterdam, the Netherlands \\
	\num 3 \addr GE HealthCare, Munich, Germany \\
    \num 4 \addr GE HealthCare, USA \\
    \num 5 \addr Brain tumor Centre, Erasmus MC Cancer Institute, Rotterdam, the Netherlands \\
    \num 6 \addr Medical Delta, Delft, the Netherlands
}

\abstract{
    Uncertainty Quantification (UQ) is a key requirement for trustworthy AI in high-stakes medical image analysis. In this work, we evaluate UQ within a multi-task Deep Learning (DL) framework for MRI-based glioma diagnosis that performs tumor segmentation and predicts Isocitrate Dehydrogenase (IDH) mutation status, 1p/19q co-deletion status and tumor grade. We use Monte Carlo Dropout (MCD) as the primary sampling-based UQ method for the detailed task-aware analysis, obtaining predictive uncertainty and its aleatoric and epistemic components. We assess uncertainty along complementary axes:  (i) MC sample convergence of uncertainty estimates and their decomposition into aleatoric and epistemic components, (ii) calibration of predictive probabilities, and (iii) operational utility of uncertainty estimates, including error detection, selective prediction, and associations with tumor segmentation performance. We also compare MCD with Deep Ensembles (DE) and Monte Carlo Deep Ensembles (MCDE) to assess whether the observed operational utility of uncertainty estimates extends beyond a single UQ method.  For the tumor segmentation task, we further examine how different voxel-wise uncertainty aggregation strategies influence case-level reliability assessment, thereby explicitly accounting for task-specific uncertainty representation. Additionally, we study task interactions to quantify how tumor segmentation quality and uncertainty relate to the prediction of the tumor features. Finally, we explore whether a composite trust score integrating tumor segmentation and classification uncertainty improves error detection. Across tasks, uncertainty estimates supported meaningful error detection, while calibration depended on the dropout rate, with moderate rates yielding the most reliable predictive probabilities. Decomposition provided task-dependent interpretability but did not consistently improve error detection over predictive uncertainty alone. The comparison with DE and MCDE showed that ensemble-based uncertainty estimates provided comparable operational utility, although no UQ method consistently dominated across all tasks and metrics. Moreover, the proposed trust score did not consistently outperform classification uncertainty for selective prediction, indicating that task-specific predictive uncertainty remains the most informative operational indicator of trust. Overall, our results provide a task-aware evaluation strategy and practical guidance towards the development of trustworthy AI for glioma diagnosis.
}

\keywords{Glioma, Magnetic Resonance Imaging, Deep Learning, Uncertainty Quantification, Monte Carlo Dropout, Deep Ensembles}

\begin{document}

\twocolumn[\maketitle]

\section{Introduction}
	\enluminure{G}{liomas} are among the most aggressive primary brain tumors and are associated with substantial morbidity and mortality worldwide \citep{ostrom2019cbtrus}. Advances in molecular neuropathology have transformed glioma classification, shifting diagnostic standards from purely histological grading towards integrated molecular definitions. The current \gls{who} classification incorporates molecular markers such as \gls{idh} mutation status and 1p/19q co-deletion status, alongside tumor grade, to stratify gliomas into biologically and clinically meaningful subtypes \citep{louis2021who}. These characteristics directly influence therapeutic strategy and patient prognosis, making accurate characterization essential for clinical decision-making.

    \gls{mri} is the primary non-invasive imaging modality used for diagnosis, treatment planning, and longitudinal monitoring of patients with glioma \citep{abdalla2020glioma}. In clinical practice, \gls{mri} is often assessed using qualitative criteria or simple measures based on tumor diameter, providing only a rudimentary view of tumor burden. Automated or semi-automated segmentation can produce more detailed and reproducible measurements of tumor subregions, supporting improved diagnosis, treatment planning, and longitudinal monitoring. However, developing robust algorithms remains challenging due to variability in tumor appearance and subtle intensity differences \citep{menze2014multimodal}. Beyond spatial delineation, \gls{dl} approaches have demonstrated promising capability in predicting molecular markers such as \gls{idh} mutation and 1p/19q co-deletion status directly from imaging data, a paradigm often referred to as radiogenomics \citep{chang2018deep, kickingereder2019radiogenomics}.

    To exploit shared imaging representations, multi-task learning frameworks have been proposed that jointly perform tumor segmentation and genetic classification within a single model architecture. Such models leverage shared feature representations and may improve efficiency and generalization compared to isolated single-task approaches \citep{kordnoori2024deep, zhou2019multitask, van2023combined}. However, while predictive performance has advanced substantially, translation of these systems into clinical workflows remains limited due to the black-box nature of \gls{nns}, which lack reliable confidence scores about their predictions \citep{lambert2024trustworthy, guo2017calibration}.

    In high-stakes medical applications, predictive accuracy alone is insufficient. AI systems must communicate the reliability of their predictions to support safe clinical integration \citep{ojha2025navigating, abdar2022need, he2025survey}. \gls{uq} has therefore emerged as a key component of trustworthy medical AI, aiming to offer rationale behind the \gls{dl} models' predictions and increase their interpretability \citep{lambert2024trustworthy}. Uncertainty estimates can enable selective automation by identifying cases requiring human review, and thus improving transparency in decision-support systems.
    
    Despite this growing interest, uncertainty in multi-task medical AI systems remains insufficiently characterized. In multi-task settings, segmentation and classification branches share intermediate representations but produce heterogeneous outputs. Uncertainty may propagate differently across tasks, and segmentation quality may influence subtype prediction reliability. Furthermore, predictive uncertainty can be decomposed into epistemic uncertainty, reflecting model uncertainty due to limited knowledge, and aleatoric uncertainty, reflecting intrinsic data ambiguity \citep{kendall2017uncertainties}. While theoretically appealing, recent work has questioned whether disentangling uncertainty sources consistently yields practical benefits in real-world settings, emphasizing that the usefulness of such decomposition for segmentation tasks may depend on dataset characteristics and downstream evaluation criteria \citep{kahl2024values}. Systematic empirical evaluation of these aspects in clinically relevant multi-task systems remains limited.
    
    These considerations highlight the need for a comprehensive and task-aware assessment of AI-based diagnostic systems. In this work, we present a systematic evaluation of \gls{uq} within a multi-task \gls{dl} framework for \gls{mri}-based glioma diagnosis. The pipeline jointly performs tumor segmentation and predicts \gls{idh} mutation status, 1p/19q co-deletion status, and tumor grade. The \gls{dl} architecture, which was initially proposed by \cite{van2023combined}, showed promising performance but lacked \gls{uq} and analysis of the interaction between the different tasks. Here, we use \gls{mcd} \citep{gal2016dropout} as the primary sampling-based \gls{uq} method to obtain predictive uncertainty and its aleatoric and epistemic components. In addition, we compare \gls{mcd} with \gls{de} and \gls{mcde} to assess whether the observed operational utility of uncertainty estimates extends beyond a single \gls{uq} method.

We evaluate uncertainty along multiple complementary axes: \gls{mc} sample convergence of uncertainty estimates, uncertainty decomposition, calibration of predictive probabilities, operational utility of uncertainty estimates including a comparison across different \gls{uq} methods, and cross-task interactions between tumor segmentation and the reliability of tumor feature prediction. We further analyze how different voxel-wise aggregation strategies affect segmentation-level uncertainty assessment. Through this comprehensive evaluation, we aim to clarify when and how uncertainty estimates meaningfully contribute to reliability in a multi-task glioma diagnosis AI system. By grounding uncertainty analysis in clinically relevant tasks and operational metrics, our study provides practical guidance towards the development of trustworthy AI for this application.
\section{Related Works}

    \subsection{\texorpdfstring{\gls{mri}}{MRI}-based radiogenomics for glioma characterization}

    Radiogenomics has emerged as a promising paradigm for the non-invasive characterization of gliomas by linking quantitative imaging features to molecular and genomic tumor profiles.  Several comprehensive reviews highlight the growing role of radiomics and radiogenomics in supporting precision medicine for glioma management, emphasizing applications ranging from molecular subtyping and risk stratification to prognosis estimation and treatment monitoring \citep{mitra2021deep, singh2021radiomics, fathi2021applications}.

     \gls{mri} constitutes one of the central imaging modalities in these efforts. Structural sequences, including \gls{t1w}, \gls{t1wce}, \gls{t2w}, and \gls{flair}, provide rich morphological information that can be exploited by radiomics pipelines for tumor characterization \citep{singh2021radiomics}. These approaches have demonstrated potential for predicting molecular subtypes, distinguishing true progression from pseudoprogression, estimating overall survival and progression-free survival, and identifying recurrence patterns \citep{fathi2021applications}.
    
    In parallel, \gls{dl}–based radiogenomics has gained substantial momentum. \gls{cnns} have been used to automatically extract hierarchical imaging features directly from \gls{mri} data, often outperforming classical radiomics approaches in molecular subtype prediction \citep{li2022molecular}. For instance, \cite{buda2020deep} demonstrated that \gls{cnns} can predict genomic subtypes of lower-grade gliomas directly from \gls{mri}, exploring both training from scratch and transfer learning strategies. Comparative studies between radiomics and \gls{dl} models suggest that learned representations may capture complementary or superior discriminative information relative to handcrafted features \citep{li2022molecular}.
    
    Despite promising performance, translation of radiogenomic models into routine clinical practice remains limited. Reviews consistently identify challenges related to reproducibility, generalizability across institutions, and robustness under distribution shifts \citep{fathi2021applications, singh2021radiomics}. In addition, uncertainty and interpretability of \gls{dl} models are increasingly recognized as critical factors for clinical acceptance \citep{singh2021radiomics}. 
    
	\subsection{Multi-task Deep Learning}
    Multi-task \gls{dl} has gained increasing attention in medical imaging as a strategy to leverage shared representations across related tasks, often leading to improved generalization, regularization, and computational efficiency compared to isolated single-task models. By jointly optimizing multiple objectives, multi-task frameworks can exploit complementary supervision signals and encourage more robust feature learning. Multi-task \gls{dl} has been successfully implemented in different medical imaging applications. For instance, \cite{zhou2019multitask} demonstrated improved breast tumor classification by combining segmentation and classification within a unified framework for ultrasound imaging.
    
    \cite{kaushik2023region} proposed a multi-task network combining segmentation and regression to generate synthetic CT images from MRI for radiation therapy planning, where
    segmentation was used to localize bone regions of interest
    and guide accurate density prediction during image translation. \cite{kordnoori2024deep} reported enhanced diagnostic performance for brain tumor characterization through joint segmentation and classification of gliomas, meningiomas, and pituitary tumors.

    Within neuro-oncology, multi-task approaches have been explored both for segmentation enhancement and for joint segmentation–classification pipelines. Several studies have focused on improving tumor delineation through auxiliary tasks. For instance, \cite{ngo2020multi} proposed a multi-task framework for small brain tumor segmentation in \gls{mri}, incorporating an auxiliary feature reconstruction task to preserve fine-grained tumor characteristics. Similarly, \cite{huang2021deep} introduced a \gls{dl} multi-task framework based on a V-Net architecture with dual decoders, where a distance transform prediction task regularized mask prediction and led to improved segmentation contour accuracy.
    
    More recent efforts have extended multi-task paradigms towards integrated diagnostic pipelines in glioma research. \cite{li2023transformer} developed a transformer-based multi-task model for simultaneous glioma segmentation and identification of infiltrated brain regions, demonstrating that shared boundary information from segmentation can enhance classification-related tasks. In a clinically oriented setting, \cite{chakrabarty2023mri} proposed a 2.5D hybrid multi-task \gls{cnn} to jointly localize, segment, and predict \gls{idh} mutation status and 1p/19q co-deletion status, integrating imaging features with prior clinical knowledge through feature fusion mechanisms.
    
   In the work of \cite{van2023combined}, a single 3D \gls{cnn} capable of jointly performing tumor segmentation and predicting \gls{idh} mutation status, 1p/19q co-deletion status, and tumor grade from pre-operative \gls{mri} was proposed. Their framework achieved good performance on an independent test set, representing one of the first unified models providing the necessary components for \gls{who} subtype derivation non-invasively. However, \gls{uq} was not incorporated, and interactions between tumor segmentation quality, reliability of tumor features prediction, and task-specific uncertainty were not systematically analyzed.

    \subsection{Uncertainty Quantification and Decomposition}

    \gls{uq} in \gls{dl} for Medical Image Analysis encompasses a broad range of approaches, including \gls{bnns}, \gls{mc} sampling methods, \gls{de}, \gls{tta}, \gls{el}, \gls{gm}, and \gls{cp} frameworks \citep{lambert2024trustworthy}. These methods differ in their ability to model aleatoric and epistemic uncertainty and in their computational requirements, particularly for high-dimensional tasks such as 3D segmentation. The same review by \cite{lambert2024trustworthy} highlights that no single method consistently dominates across clinical applications, and emphasizes the importance of task-specific and application-driven evaluation of uncertainty estimates. \gls{mcd} remained the most used method for \gls{uq} among 218 papers analyzed in this review, followed by \gls{de}.

    Several approaches have been proposed to decompose predictive uncertainty into aleatoric (data-related) and epistemic (model-related) components. In a Bayesian framework, the components are obtained via Predictive Entropy and \gls{mi} \citep{smith2018understanding, mukhoti2023deep}. More recent methods derive uncertainty components directly from network outputs, for example within 3D U-Net architectures \citep{jones2022direct} or single-model diffusion-based frameworks \citep{chan2024estimating}, demonstrating potential clinical relevance for identifying ambiguous or out-of-distribution cases.


    However, the practical value of disentanglement remains contested. \cite{kahl2024values} showed that successful separation in controlled settings for segmentation tasks does not necessarily translate to real-world medical data, and that its downstream benefit strongly depends on the task, dataset properties, and aggregation strategy.  Similarly, \cite{mukhoti2023deep} emphasized that predictive entropy summarizes total uncertainty and therefore confounds aleatoric and epistemic uncertainty. They showed that predictive entropy can still be effective when one uncertainty source is low or dominant, but may become less informative when ambiguous in-distribution samples and out-of-distribution samples both lead to high predictive entropy. These findings suggest that epistemic--aleatoric decomposition may be useful for interpreting the source of uncertainty, but that its operational benefit should not be assumed and instead needs to be validated within the specific application context.
    
    \subsection{Uncertainty Quantification in Multi-Task Settings}

    While \gls{uq} has been extensively studied for single-task segmentation or classification, its role in multi-task learning remains comparatively underexplored. In particular, few works explicitly analyze how uncertainty behaves across interacting tasks or how it propagates within shared representations.

      \cite{ruan2020mt} proposed Mt-UcGAN, a unified framework for joint tumor segmentation, quantification, and uncertainty estimation in renal CT imaging, incorporating an uncertainty-constrained adversarial mechanism within a multi-task architecture. Similarly, \cite{mehta2021propagating} investigated the propagation of voxel-level uncertainty across cascaded tasks, demonstrating that uncertainty estimates from one stage can improve performance in downstream segmentation tasks. However, these studies focus primarily on segmentation or cascaded processing pipelines rather than on tightly integrated frameworks that jointly perform different tasks.
    
    To the best of our knowledge, the interaction between tumor segmentation uncertainty and molecular subtype prediction uncertainty has not been systematically investigated in multi-task AI frameworks focused on glioma diagnosis. Although segmentation has been used in multi-task approaches under the theoretical assumption that tumor shape information can guide AI models in learning relevant features for the tumor subtyping, it remains unclear whether uncertainty in the shared segmentation backbone influences reliability in the prediction of relevant tumor features, and whether cross-task aggregation improves operational trust. 
\section{Methods}

\subsection{Multi-task Deep Learning Glioma Subtyping and Uncertainty Quantification Framework}

\begin{figure*}[t]
  \centering
  \includegraphics[width=1\linewidth, height=0.4\textheight]{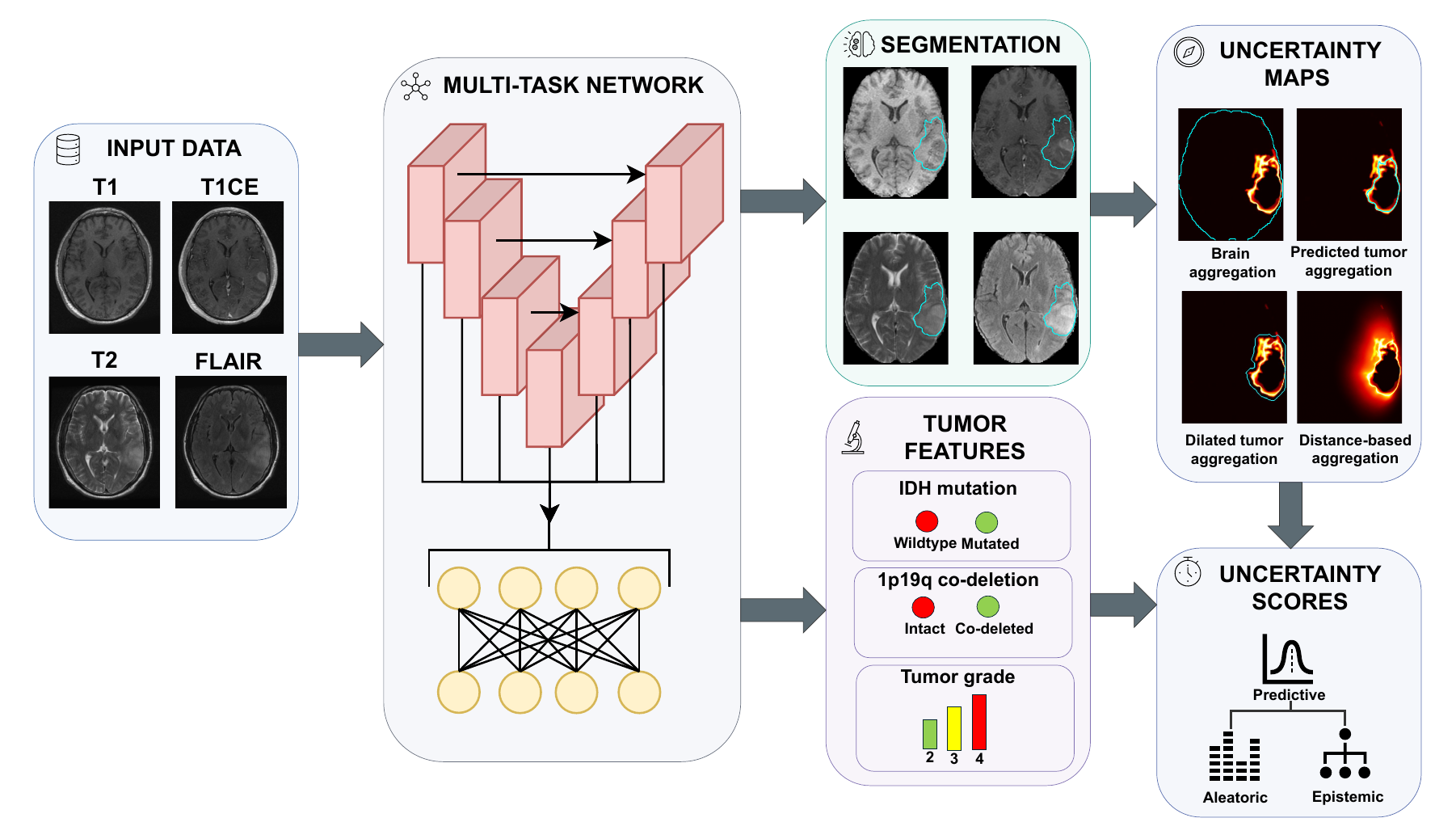}
  \caption{Overview of the multi-task glioma subtyping framework with uncertainty quantification. The input data consist of four structural MRI sequences per patient, namely T1-weighted (T1w), contrast-enhanced T1-weighted (T1wCE), T2-weighted (T2w), and fluid-attenuated inversion recovery (FLAIR). The sequences are preprocessed using registration, bias field correction, and skull stripping. Subsequently, the data are used to train a Deep Learning model that jointly performs tumor segmentation and prediction of IDH mutation status, 1p/19q co-deletion status, and tumor grade. For each task, predictive, aleatoric, and epistemic uncertainty are computed. For tumor segmentation, voxel-wise uncertainty maps are converted into case-level uncertainty scores using four aggregation strategies: brain-level aggregation, predicted-tumor aggregation, dilated-tumor aggregation, and distance-weighted aggregation around the tumor region. In the tumor features panel, the colors of the icons vary from tumors associated with poorer prognosis (red) to those with better prognosis (green). }
  \label{fig:method_overview}
\end{figure*}

As shown in Figure \ref{fig:method_overview}, the pipeline operates on four pre-operative structural \gls{mri} sequences: \gls{t1w}, \gls{t1wce}, \gls{t2w}, and \gls{flair}. The end goal is to accurately predict three key glioma characteristics: \gls{idh} mutation status (wildtype or mutated), 1p/19q co-deletion status (intact or co-deleted), and tumor grade (grade 2, 3, or 4).  

The framework includes a pre-processing module, and an encoder-decoder network \citep{van2023combined, ronneberger2015u} that performs tumor segmentation as an auxiliary task to extract imaging features at different resolution levels both in the encoder and decoder pathways. These features are concatenated and fed to a classification branch composed of fully connected layers for the prediction of each of the tumor features. Details on the pre-processing and architecture can be found in Appendix \ref{app:pre-processing} and \ref{app:architecture}, respectively.

\subsection{Data}

In compliance with the expected input of the model, the dataset was composed of different subsets containing the four structural \gls{mri} sequences, accompanied by its corresponding tumor segmentation and labels (when available), regarding \gls{idh} mutation status, 1p/19q co-deletion status and tumor grade.

The train and test sets are comprised of a collection of both in-house and publicly available datasets of patients with adult-type glioma, including BraTS \citep{bakas2017advancing,bakas2018identifying,menze2014multimodal}, Brain tumor Progression \citep{schmainda2018data}, CPTAC-GBM \citep{cptac}, Erasmus Glioma Database (EGD) \citep{van2021erasmus}, IvyGAP \citep{shah2016data} and REMBRANDT \citep{scarpace2019data} for the training set; and TCGA-GBM \citep{scarpace2016cancer} and TCGA-LGG \citep{pedano2016cancer} for the test set. The distribution of the data is presented in Table \ref{tab:datasets-general-distribution}.

\begin{table}[h]
\centering
\caption{Train and test data distribution}
\label{tab:datasets-general-distribution}
\begin{tabular}{@{}cccc@{}}
\toprule
Set & Dataset & Number of cases & Total \\ 
\midrule
\multirow{7}{*}{Train} 
& BraTS     & 156 & \multirow{7}{*}{1466} \\        
& \shortstack{Brain tumor\\Progression} & 20 \\
& CPTAC-GBM & 45  &                       \\
& EGD       & 775 &                       \\
& IvyGAP    & 39  &                       \\
& In-house  & 322 &                       \\
& REMBRANDT & 109 &                       \\ 
\midrule
\multirow{2}{*}{Test}  
& TCGA-GBM  & 133 & \multirow{2}{*}{236}  \\
& TCGA-LGG  & 103 &                       \\ 
\bottomrule
\end{tabular}
\end{table}

Since not all data samples had information regarding molecular features and tumor grade, this was handled with a masked loss during training. Details on this procedure are given in Appendix \ref{app:architecture}. Table \ref{tab:data-detailed-overview} in Appendix \ref{ap2:data} provides additional information on the data distribution for each of the tumor features, evidencing the highly-imbalanced nature of all the tasks across datasets.

\subsection{Uncertainty Quantification}

\subsubsection{Bayesian Modeling}

Given a training dataset consisting of input
$\mathcal{X} = \{x_1, \dots, x_n\}$ 
and corresponding output
$\mathcal{Y} = \{y_1, \dots, y_n\}$, 
Bayesian modeling aims to infer parameters $\boldsymbol{\omega}$ of a function 
$y = f^{\boldsymbol{\omega}}(x)$ that are likely to have generated the observed output.

Before observing any data, an initial belief about plausible parameter values is set. 
In the Bayesian framework, this belief is encoded through a prior distribution over the parameter space, denoted as $p(\boldsymbol{\omega})$. 
The prior reflects assumptions about the model parameters before incorporating evidence from the data.

Once data are observed, the prior belief is updated. 
To perform this update a likelihood function is required. This function specifies the probabilistic mechanism by which outputs are generated given inputs and parameters, thus describing how well a particular parameter configuration explains the observed data. It is defined as:

\begin{equation}
p(\mathcal{Y} \mid \mathcal{X}, \boldsymbol{\omega}).
\end{equation}

Applying Bayes’ theorem, the posterior distribution over the parameters is given by

\begin{equation}
p(\boldsymbol{\omega} \mid \mathcal{X}, \mathcal{Y}) 
= \frac{p(\mathcal{Y} \mid \mathcal{X}, \boldsymbol{\omega}) \, p(\boldsymbol{\omega})}
{p(\mathcal{Y} \mid \mathcal{X})},
\end{equation}

\noindent
where $p(\mathcal{Y} \mid \mathcal{X})$ is the marginal likelihood (or evidence), 
and acts as a normalization constant ensuring that the posterior integrates to one.

The posterior distribution characterizes the updated belief over model parameters after observing the data. 
Given a new input $x^{*}$, predictions are obtained by marginalizing over all possible parameter configurations, weighted by their posterior probability:

\begin{equation}
\label{eq:pred-distri}
p(y^{*} \mid x^{*}, \mathcal{X}, \mathcal{Y}) 
= \int p(y^{*} \mid x^{*}, \boldsymbol{\omega}) \,
p(\boldsymbol{\omega} \mid \mathcal{X}, \mathcal{Y}) \, d\boldsymbol{\omega}.
\end{equation}

This integral expresses the Bayesian predictive distribution, which accounts for uncertainty in the model parameters by averaging predictions over the posterior distribution.

\subsubsection{Uncertainty Decomposition and MCD}
\label{subsec:uncertainty-decomposition}
Let $N$ be a \gls{nn} with weights $\mathbf{W}$. The theoretical definition of the predictive distribution in Equation \ref{eq:pred-distri} is analytically intractable for $N$ given its high-dimensional weight space.
\gls{mcd} \citep{gal2016dropout} was then proposed as an approximation to Bayesian inference.

 In \gls{mcd}, Bayesian inference is approximated by applying dropout to $N$ with rate $\delta$ at train time, and then performing $T$ stochastic forward passes with the same dropout rate enabled at test time. Given some training data $\mathcal{D}$ and a new input $x^*$, each forward pass $t$ samples a different set of network weights \(\mathbf{W}_t\), yielding a collection of predictions \(\{p(y^* | x^*, \mathbf{W}_t)\}_{t=1}^T\). The predictive distribution is then approximated by:

\begin{equation}
p(y^* | x^*, \mathcal{D}) \approx \frac{1}{T} \sum_{t=1}^T p(y^* | x^*, \mathbf{W}_t).
\end{equation}

\noindent
The total predictive uncertainty can be quantified by computing the entropy of this predictive distribution \citep{smith2018understanding,depeweg2018decomposition}:

\begin{align}
\label{eq:pred-unc}
U_{\text{predictive}}
&= - \sum_{c}
\left(
\frac{1}{T} \sum_{t=1}^{T} p(y^* = c \mid x^*, \mathbf{W}_t)
\right) \notag \\
&\quad \times
\log \left(
\frac{1}{T} \sum_{t=1}^{T} p(y^* = c \mid x^*, \mathbf{W}_t)
\right),
\end{align}

\noindent
where $x^*$ denotes a test input sample and $y^*$ the corresponding predicted label. 
$\mathbf{W}_t$ represents the network weights obtained at \gls{mc} 
sample $t$ when dropout is applied at test time, and $T$ is the total number 
of \gls{mc} forward passes (\gls{mc} samples hereinafter). The term $p(y^* = c \mid x^*, \mathbf{W}_t)$ denotes 
the predicted probability for class $c$ produced by the model in the $t$-th \gls{mc} sample. The inner average $\frac{1}{T}\sum_{t=1}^{T} p(y^* = c \mid x^*, \mathbf{W}_t)$ 
approximates the predictive distribution by averaging the class probabilities across 
all \gls{mc} samples. Finally, $c$ indexes the set of possible classes.

The predictive uncertainty can be decomposed into two components. The aleatoric uncertainty, representing the inherent noise in the data, is estimated by averaging the entropy of the predictive distribution from all $T$ \gls{mc} samples \citep{smith2018understanding,depeweg2018decomposition}:

\begin{align}
\label{eq:al-unc}
U_{\text{aleatoric}}
&= \frac{1}{T} \sum_{t=1}^T 
\Bigg(
- \sum_{c} 
p(y^* = c \mid x^*, \mathbf{W}_t) \notag \\
&\quad \times \log p(y^* = c \mid x^*, \mathbf{W}_t)
\Bigg),
\end{align}

\noindent
where $p(y^* = c \mid x^*, \mathbf{W}_t)$ represents the predicted probability for 
class $c$ at inference under weight realization $\mathbf{W}_t$. In this case, 
the entropy is computed for each \gls{mc} sample $t$ and subsequently averaged across the $T$ samples.

The epistemic uncertainty describes the uncertainty coming from the model parameters. It is computed as the difference between the predictive and aleatoric components, yielding the \gls{mi} \citep{smith2018understanding,depeweg2018decomposition}:

\begin{equation}
\label{eq:ep-unc}
U_{\text{epistemic}} =  
 \left( \frac{1}{T} \sum_{t=1}^{T} \sum_{c} p_{t,c} \log p_{t,c} \right) - \left(  \sum_{c} \bar{p}_c \log \bar{p}_c \right),
\end{equation}

\noindent where $\bar{p}_c$ represents the mean predicted probability for class $c$ across all $T$ \gls{mc} samples, and $p_{t,c}$ describes the class $c$ probability at \gls{mc} sample $t$.

\subsection{Evaluation Experiments:} 

\subsubsection{MC Sample Convergence and Decomposition of Uncertainty Estimates}

We performed a systematic evaluation of \gls{mcd} parameters, varying the dropout rate $\delta$ and the number of \gls{mc} samples $T$ and observing the magnitudes for predictive, aleatoric and epistemic uncertainty, which were computed as shown in equations \ref{eq:pred-unc}, \ref{eq:al-unc} and \ref{eq:ep-unc}. Specifically, we explored values of $\delta$ between 0.2 and 0.6 in increments of 0.05, and values of $T$ from 10 to 100 in steps of 10. These ranges were selected to analyze the effect of both parameters in a wide spectrum, covering both moderate and higher values, and considering theoretical and empirical insights from the literature \citep{gal2016dropout,kendall2015bayesian,srivastava2014dropout}.

We define the dropout rate as the probability of dropping out units, meaning that higher rates reduce the network predictive capacity more aggressively. Too low a dropout rate may lead to underestimation of uncertainty, while excessively high values can degrade predictive performance or lead to underfitting. The upper bound of 0.6 is chosen such that even with aggressive dropout, sufficient model capacity is retained without requiring network width scaling. Likewise, the number of \gls{mc} samples influences the fidelity of uncertainty estimates: too few may yield noisy approximations, but the more samples, the longer the computation time at inference.

A sufficient number of \gls{mc} samples was defined as the point at which the average uncertainty estimates showed limited additional change with increasing sampling.

\subsubsection{Calibration and Quality Analysis}
\label{subsection:calibration}

Model calibration in the classification tasks was assessed using \gls{ece} and \gls{nll}. \gls{ece} quantifies the discrepancy between predicted confidence and empirical accuracy. Given predicted labels $\hat{y}_i$ and predicted confidence values $\hat{p}_i$ for a case $i$ from the total number of cases $N$, predictions are partitioned into $B$ confidence bins. Here, $\hat{p}_i$ denotes the maximum softmax probability assigned to the predicted class $\hat{y}_i$. For bin $b$, let $\mathrm{acc}(b)$ denote the empirical accuracy and $\mathrm{conf}(b)$ the mean confidence. \gls{ece} is computed as:

\begin{equation}
\mathrm{\gls{ece}} = \sum_{b=1}^{B} \frac{|B_b|}{N} \left| \mathrm{acc}(b) - \mathrm{conf}(b) \right|,
\end{equation}

\noindent
where $|B_b|$ is the number of samples in bin $b$. We used $B=5$ equally spaced confidence bins across all classification tasks and dropout configurations. This was a practical choice to balance calibration-curve resolution with the reliability of bin-wise accuracy estimates, since using more bins reduces the number of cases per bin and can make \gls{ece} more sensitive to small changes in case assignment.

\gls{nll} evaluates both correctness and confidence by penalizing low probability assigned to the true class:

\begin{equation}
\mathrm{\gls{nll}} = - \frac{1}{N} \sum_{i=1}^{N} \log \hat{p}_{i,y_i},
\end{equation}

\noindent
where $\hat{p}_{i,y_i}$ denotes the predicted probability assigned to the true class $y_i$ for case $i$.

Importantly, both \gls{ece} and \gls{nll} were computed exclusively from the softmax output probabilities of the \gls{mcd} ensemble and therefore reflect calibration of the model probability outputs, rather than calibration of the uncertainty estimates themselves.

Additionally, we computed the \gls{auc} of the \gls{mcd} ensemble model at each configuration as a general measure of classification performance using the softmax probabilities. In case of the tumor grade task, which consisted of three classes, it was calculated using the one-vs-rest strategy.

For the tumor segmentation task, the quality of uncertainty estimates was assessed by computing the Pearson correlation coefficient $\rho$ between case-level uncertainty scores and the \gls{dsc} between the predicted segmentation and the ground truth for all cases in the test set, under the assumption that informative uncertainty estimates should correlate negatively with tumor segmentation accuracy. 
To assess whether the observed relationships were robust to non-linear but monotonic trends, we additionally computed Spearman rank correlation $\rho_{s }$.

Case-level uncertainty is derived by aggregating entropy-based voxel-wise uncertainty estimates.  This aggregation was explored in four different regions. First, full-brain aggregation was computed by averaging uncertainty over the brain mask obtained using HD-BET \citep{isensee2019automated}. Second, predicted tumor aggregation was computed by averaging uncertainty within the predicted tumor mask obtained from the \gls{mcd} ensemble. Third, to include uncertainty immediately adjacent to the predicted tumor boundary, we computed a dilated predicted-tumor aggregation, where the predicted tumor mask was expanded using a 5-voxel binary dilation. Finally, we evaluated a boundary-weighted aggregation in which all brain voxels contributed to the case-level score with weights decreasing as a function of their distance to the predicted tumor boundary. Specifically, for voxel $v$, the weight was defined as
\[
w(v) = \exp\left(-\frac{d(v)^2}{2\sigma^2}\right),
\]
where $d(v)$ is the Euclidean distance from voxel $v$ to the predicted tumor boundary for a given case, and $\sigma=5$ voxels. The boundary-weighted uncertainty score was then computed as
\[
U^{\mathrm{bw}} =
\frac{\sum_{v \in \Omega^{\mathrm{brain}}} w(v) u_v}
{\sum_{v \in \Omega^{\mathrm{brain}}} w(v)},
\]
with $u_v$ denoting the uncertainty value at voxel $v$. This allowed us to compare whole-brain, tumor-local, peri-tumor, and boundary-aware aggregation strategies.

\subsubsection{Operational Utility of Uncertainty Estimates}
\label{sec:operational-utility}

To evaluate the operational utility of uncertainty estimates, we quantified their ability to identify erroneous predictions across tasks and to support selective prediction by retaining the most certain cases. Table~\ref{tab:uncertainty_metrics_summary} provides a summary of the metrics used, their interpretation, and expected ranges indicative of strong performance. 

\begin{table*}[h]
\centering
\caption{Summary of the metrics used to assess operational utility of uncertainty estimates, namely, Uncertainty-based Receiver Operating Characteristic Area Under the Curve (U-AUC), Average Precision (AP), Lift, and Area Under the Risk--Coverage Curve (AURC). For each metric, the purpose, possible values, and interpretation are presented.}
\label{tab:uncertainty_metrics_summary}
\begin{tabular}{llll}
\hline
\multicolumn{1}{c}{Metric} & \multicolumn{1}{c}{Purpose} & \multicolumn{1}{c}{\begin{tabular}[c]{@{}c@{}}Possible \\ Values\end{tabular}} & \multicolumn{1}{c}{Interpretation} \\ \hline
U-AUC & \begin{tabular}[c]{@{}l@{}}Measures the probability that a \\ randomly chosen error case has \\ higher uncertainty than a correct \\ case.\end{tabular} & \multicolumn{1}{c}{0 -- 1} & \begin{tabular}[c]{@{}l@{}}Values close to 1 indicate\\ uncertainty correctly ranks \\ errors above successes.\end{tabular} \\
AP & \begin{tabular}[c]{@{}l@{}}Evaluates precision-recall \\ trade-off for predicting errors, \\ suitable for imbalanced tasks.\end{tabular} & \multicolumn{1}{c}{0 -- 1} & \begin{tabular}[c]{@{}l@{}}Values above the baseline \\ error rate ($\epsilon$) indicate better-than-random \\ error identification.\end{tabular} \\
Lift & \begin{tabular}[c]{@{}l@{}}Quantifies error-identification\\ performance relative to baseline \\ prevalence.\end{tabular} & \multicolumn{1}{c}{\(0 - \dfrac{1}{\epsilon}\)} & \begin{tabular}[c]{@{}l@{}}Values $>1$ indicate that uncertainty \\ successfully prioritizes error cases.\\ A value of 1 indicates random \\ selection.\end{tabular} \\
 AURC & \begin{tabular}[c]{@{}l@{}}Evaluates the mean residual risk\\ among retained cases as coverage\\ varies from most certain to all cases.\end{tabular} & \multicolumn{1}{c}{0 -- 1} & \begin{tabular}[c]{@{}l@{}}Lower values indicate that\\ uncertainty successfully retains\\ lower-risk cases at reduced coverage.\end{tabular} \\ \hline
\end{tabular}
\end{table*}

For each classification task, a binary error indicator for case $i$ was defined as:
\begin{equation}
e_{i} =
\begin{cases}
1, & \text{if prediction } i \text{ is incorrect} \\
0, & \text{otherwise}.
\end{cases}
\end{equation}

For the tumor segmentation, the binary error indicator was defined as:
\begin{equation}
e_{i}^{\mathrm{seg}} =
\begin{cases}
1, & \text{if ${DSC}_i \leq \tau$}\\
0, & \text{otherwise}.
\end{cases}
\end{equation}

\noindent Since this binary definition of segmentation error depends on the selected \gls{dsc} threshold $\tau$, we additionally performed a threshold-sensitivity analysis for $\tau \in \{0.60, 0.70, 0.80\}$. We computed the segmentation error rate to assess whether the operational utility of segmentation uncertainty depended heavily on the value of $\tau$, or whether the conclusions remained consistent across more lenient and stricter definitions of error.


\textit{U-AUC:} In this case, we measured the \gls{u-auc}: \gls{mcd}-derived uncertainty scores $U$ were treated as continuous predictors of error $e_{i}$, and the area under this curve was computed. In this setting, this metric evaluates the probability that a randomly chosen erroneous case receives a higher uncertainty score than a randomly chosen correct case:
\begin{equation}
 U\text{-}AUC = \mathbb{P}(U_{e=1} > U_{e=0}).
\end{equation}

\noindent While \gls{u-auc} is threshold-independent, it may be optimistic when errors are rare. It aims to answer the question: \textit{does uncertainty rank errors above successes overall?}

\textit{\gls{ap}:} Given the typically imbalanced nature of medical imaging tasks, we additionally computed the \gls{ap}, corresponding to the area under the precision-recall curve:
\begin{equation}
\mathrm{AP} = \sum_{k} (R_k - R_{k-1}) P_k,
\end{equation}
where $P_k$ and $R_k$ denote precision and recall at threshold $k$.

The baseline \gls{ap} of a random classifier equals the error rate:
\begin{equation}
\epsilon = \frac{1}{N} \sum_{i=1}^{N} e_i.
\end{equation}

Thus, \gls{ap} values above $\epsilon$ indicate better-than-random identification of errors. It aims to answer the question: \textit{Is uncertainty useful when errors are rare?}

\textit{Error-Rate Lift:} To quantify error identification relative to baseline prevalence, we computed Lift:
\begin{equation}
\mathrm{Lift} = \frac{\mathrm{AP}}{\epsilon}.
\end{equation}

A Lift of 1 indicates performance equivalent to the baseline error rate, as expected under random ranking. A Lift greater than 1 indicates that erroneous cases tend to receive higher uncertainty scores than non-erroneous cases.

\textit{\gls{aurc}:} To additionally evaluate uncertainty estimates in a selective prediction setting, we computed the \gls{aurc} \citep{el2010foundations,jaeger2022call,zenk2025comparative}. 
For our application, we computed an empirical definition obtained by progressively retaining cases from lowest to highest uncertainty. Let $U_i$ denote the uncertainty score for case $i$, with larger values indicating lower confidence. The $N$ cases were ordered from most certain to least certain, such that $U_{(1)} \leq U_{(2)} \leq \dots \leq U_{(N)}$. The case-wise risk $r_i$ was defined as:
\begin{equation}
r_i =
\begin{cases}
e_i, & \text{for classification tasks},\\
1-\mathrm{DSC}_i, & \text{for tumor segmentation},
\end{cases}
\end{equation}
where $e_i=1$ for an incorrect classification and $e_i=0$ otherwise.

For a retained set containing the $k$ most certain cases, coverage was defined as:
\begin{equation}
C_k = \frac{k}{N},
\end{equation}
and selective risk was computed as the mean risk among the retained cases:
\begin{equation}
R(C_k) = \frac{1}{k}\sum_{j=1}^{k} r_{(j)}.
\end{equation}

The \gls{aurc} was then computed by averaging the selective risk over all retained-set sizes:
\begin{equation}
\mathrm{AURC}
=
\frac{1}{N}\sum_{k=1}^{N} R(C_k).
\end{equation}
This corresponds to a discrete retained-set approximation of the risk--coverage curve area. Lower \gls{aurc} values indicate that low-uncertainty retained cases have lower residual risk across coverage levels. This metric therefore aims to answer the question: \textit{how much prediction risk remains when retaining only the most certain cases?}

\subsubsection{Comparison of MCD with other UQ methods}
\label{sec:ensemble-uq-comparison}



Although \gls{mcd} is widely used for uncertainty estimation in \gls{dl}, relying on a single \gls{uq} method may limit the interpretation of a task-aware uncertainty analysis. After selecting the final \gls{mcd} configuration, we therefore compared it with two ensemble-based strategies, \gls{de} and \gls{mcde}. These methods were selected because they are compatible with the multi-task architecture used in this study. Ensembles can also be interpreted as Bayesian Model Averaging \citep {he2020bayesian,wilson2020bayesian,mukhoti2023deep}. Therefore, the same framework for disentangling epistemic and aleatoric uncertainty described in \ref{subsec:uncertainty-decomposition} was applied.

For \gls{de}, we trained $M=5$ independently initialized models using the same architecture, data split, preprocessing, loss functions, and training procedure (including dropout) as in the main experiments. The ensemble members differed only in their fixed random seeds, which affected initialization, mini-batch ordering, and stochastic data augmentation. At inference, dropout was disabled and the final predictive probability vector was obtained by averaging the deterministic probability vectors across ensemble members:

\[
\bar{\mathbf{p}}^{\mathrm{DE}}_i =
\frac{1}{M}\sum_{m=1}^{M}\mathbf{p}_{i,m},
\]

where $\mathbf{p}_{i,m}$ denotes the class-probability vector for case $i$ predicted by ensemble member $m$.

For \gls{mcde}, we combined deep ensembles with \gls{mcd} by enabling dropout at inference for each independently trained ensemble member. For each model, we used the selected \gls{mcd} configuration identified in the \gls{mcd} analyses. The final \gls{mcde} predictive probability vector was obtained by averaging predictions across both ensemble members and \gls{mc} samples:

\[
\bar{\mathbf{p}}^{\mathrm{MCDE}}_i =
\frac{1}{MT}\sum_{m=1}^{M}\sum_{t=1}^{T}\mathbf{p}_{i,m,t},
\]

where $\mathbf{p}_{i,m,t}$ denotes the probability vector for case $i$ from the $t$-th \gls{mc} sample of ensemble member $m$.


This additional comparison was included to provide a broader evaluation of \gls{uq} beyond \gls{mcd} and to assess whether the operational utility of uncertainty estimates was consistent across dropout-based and ensemble-based sampling strategies. It was performed after selecting the final \gls{mcd} configuration and using the same operational-utility metrics defined in Section~\ref{sec:operational-utility}: \gls{u-auc}, \gls{ap}, Lift, and \gls{aurc}. This allowed us to assess whether the operational utility observed for \gls{mcd} was specific to dropout-based stochastic inference, or whether comparable behavior was also observed for ensemble-based uncertainty estimates.

\subsubsection{Interaction Between Tumor Segmentation and Classification}

To investigate cross-task dependencies, we analyzed the relationship between tumor segmentation performance and uncertainty, as well as classification correctness.

\textit{Tumor Segmentation Quality and Classification Correctness:}
We used \gls{dsc} scores as a measure of tumor segmentation quality and compared its values between correctly and incorrectly classified cases for each of the tasks. Differences between groups were evaluated using the Mann–Whitney U test to assess whether segmentation performance systematically differed between classification outcomes.

\textit{Uncertainty Correlation Across Tasks:} To assess shared uncertainty patterns, we computed Pearson correlation coefficients between tumor segmentation uncertainty $U_{\text{seg}}$ and classification uncertainty $U_{\text{cls}}$. This analysis was performed separately for predictive, aleatoric, and epistemic components.

\textit{Tumor Segmentation Uncertainty as Predictor of Classification Error:}
Finally, we evaluated whether segmentation uncertainty alone was able to predict classification errors by computing \gls{u-auc}, \gls{ap}, Lift and \gls{aurc} using $U_{\text{seg}}$ as the predictor and $e_{\text{cls}}$ as the outcome. This analysis quantified whether uncertainty in the shared segmentation backbone propagated to prediction reliability of the tumor features.

\textit{Composite Trust Score:}
\label{subsec:composite-trust-score}

To investigate whether incorporating tumor segmentation uncertainty improved the identification of classification errors beyond classification uncertainty alone, we defined a composite trust score that integrated uncertainty from both tasks. The trust score for task 
$t$ and case $i$ was defined as:

\begin{equation}
\mathrm{Trust}_{i,t} = 1 - \left( \alpha_t \, \tilde{U}_{i,t} + (1 - \alpha_t) \, \tilde{U}_{i,\text{seg}} \right),
\end{equation}

\noindent
where $\tilde{U}_{i,t}$ and $\tilde{U}_{i,\text{seg}}$ denote the normalized predictive uncertainty for the classification and the segmentation task, respectively, and $\alpha_t \in [0,1]$ represents the weighting factor for task $t$.

The weight $\alpha_t$ was determined based on the relative ability of each uncertainty source to identify classification errors, quantified using Lift. Specifically, let $\mathrm{Lift}_t$ denote the Lift obtained when using classification uncertainty to predict classification errors on task $t$, and $\mathrm{Lift}_{\text{seg} \rightarrow t}$ the Lift obtained when using tumor segmentation uncertainty for the same purpose. The task-specific weight was defined as:

\begin{equation}
\alpha_t =
\frac{\mathrm{Lift}_t}
{\mathrm{Lift}_t + \mathrm{Lift}_{\text{seg} \rightarrow t}}.
\end{equation}
This formulation assigns greater weight to the uncertainty source with greater error-identification performance relative to baseline prevalence. Intuitively, if classification uncertainty is highly predictive of errors, $\alpha_t$ approaches 1 and the trust score relies primarily on classification uncertainty. Conversely, if segmentation uncertainty provides additional predictive value, its contribution increases accordingly.

The subtraction from 1 ensures that higher trust scores correspond to more reliable predictions, facilitating interpretation and enabling direct comparison with confidence-based referral strategies. The proposed trust score was evaluated using selective prediction analysis to determine whether combining uncertainty sources improved error identification compared to classification uncertainty alone.

\section{Results}
\label{sec:results}
\subsection{MC-Sample Convergence and Decomposition of Uncertainty Estimates}

\begin{figure*}[t]
		\centering
		\includegraphics[width=1\linewidth, height=0.5\textheight]{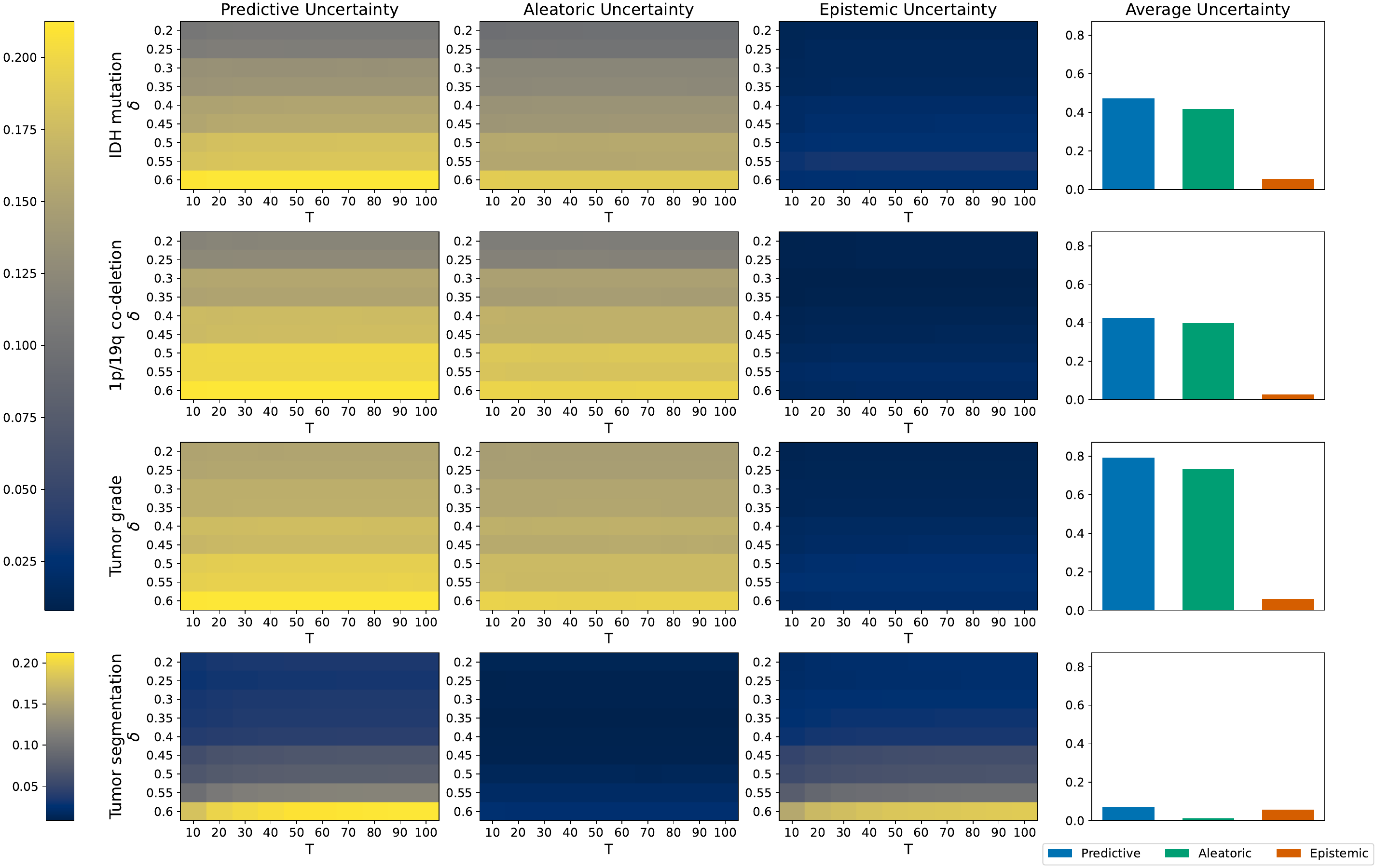}
		  \caption{First three columns from left to right: predictive, aleatoric and epistemic uncertainty value heatmaps across all tasks as a function of the number of \gls{mcd} samples $T$ and the dropout rate $\delta$. The value at each cell of the heatmap represents the mean uncertainty across all cases of the test set. The last column presents the average value for each type of uncertainty over all experiments with the blue, green and red bars representing predictive, aleatoric and epistemic uncertainty, respectively. Uncertainty values for classification tasks were normalized at the same scale to ease the analysis of the patterns. For the segmentation task the heatmaps were kept in the original scale.}
        \label{fig:effect_mcd}
	\end{figure*}

As shown in Figure \ref{fig:effect_mcd}, the average uncertainty values reached a plateau across tasks once the number of MC samples was approximately 20--30, depending on the task.
 Increasing dropout rates generally led to proportional increases in predictive, aleatoric and epistemic uncertainties.

For all the tumor features prediction tasks, namely, \gls{idh} mutation status, 1p/19q co-deletion status, and tumor grade, the aleatoric uncertainty component was the main contributor to the predictive uncertainty. Conversely, epistemic uncertainty dominated in the tumor segmentation task.

\subsection{Calibration and Quality Analysis}

\begin{figure*}[t]
    \centering
    \includegraphics[width=1\linewidth, height=0.5\textheight]{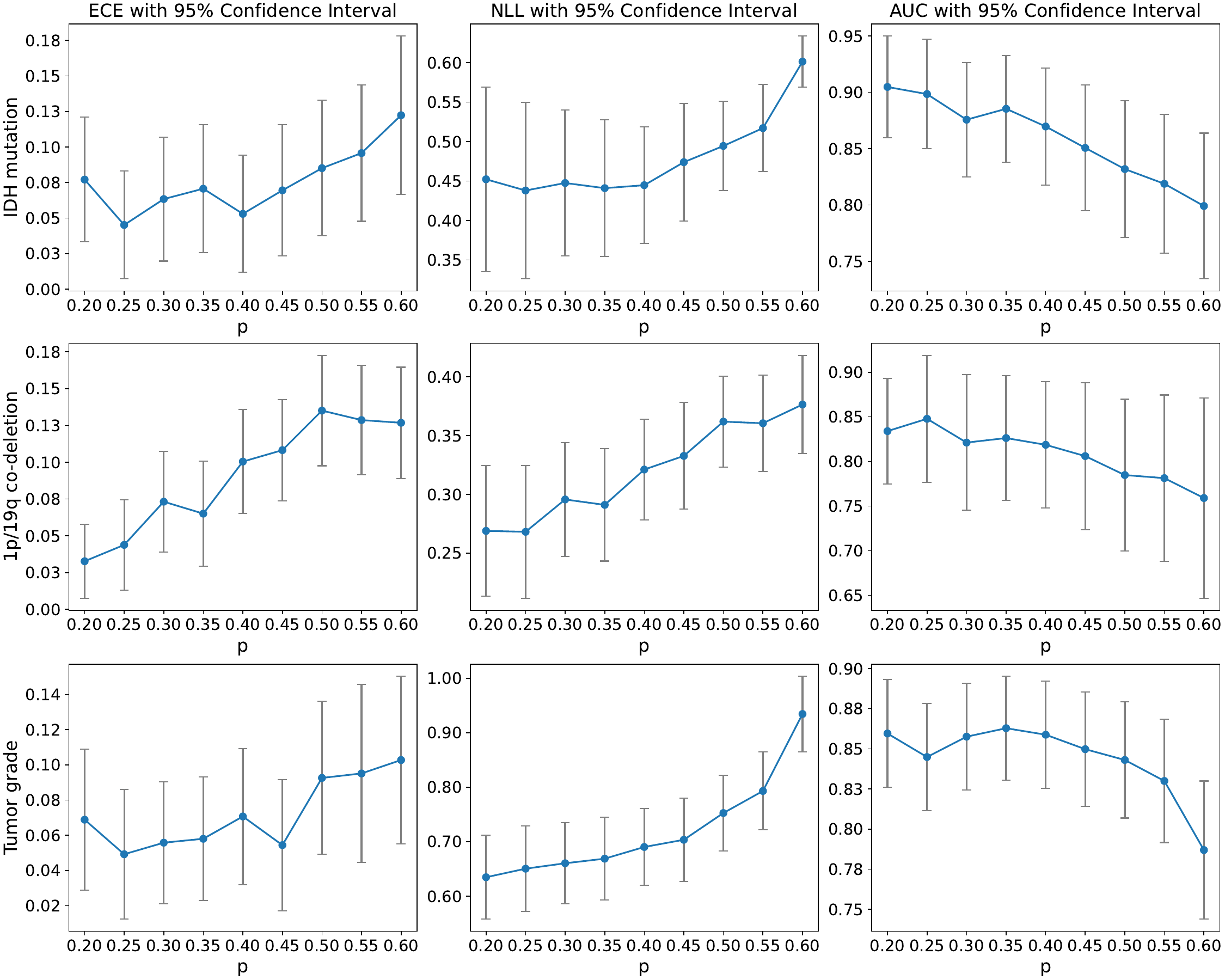}
      \caption{Expected Calibration Error (ECE), Negative Log-Likelihood (NLL) and Receiver
    Operating Characteristic Area Under the Curve (AUC) values as a function of dropout rate $\delta$. For each point in every graph, a $95\%$ confidence interval is presented. This interval was obtained by $1000\times$ bootstrap resampling of the test set. }
  \label{fig:calibration_analysis}
\end{figure*}

Based on the convergence patterns in Figure \ref{fig:effect_mcd}, we selected $T=30$ for the remaining analyses, as this was the point at which average uncertainty values had generally plateaued across tasks.
 Figure \ref{fig:calibration_analysis} presents the \gls{ece}, \gls{nll} and \gls{auc} as a function of dropout rate $\delta$ for the classification tasks. For the \gls{idh} mutation status prediction, both \gls{ece} and \gls{nll} reached their lowest values at dropout 0.25, with subtle fluctuations between 0.2 and 0.4 and then a steady increase thereafter. \gls{auc} was highest at 0.2 and generally declined as dropout increased. In the case of 1p/19q co-deletion status prediction, \gls{ece} and \gls{nll} were minimized at 0.2 and generally rose with higher dropout rates. \gls{auc} followed a similar decreasing pattern. For the tumor grade prediction, \gls{ece} was lowest at 0.25 and gradually increased with dropout, while \gls{nll} rose consistently. \gls{auc} also showed a decreasing trend for dropout rates higher than 0.35.

Although \gls{ece}, \gls{nll}, and \gls{auc} showed broadly related trends, the curves were not identical across dropout rates. For \gls{idh} mutation status prediction, \gls{ece} showed a local decrease between dropout rates 0.35 and 0.40, whereas \gls{nll} did not show a corresponding decrease and \gls{auc} remained below its value at lower dropout rates. For 1p/19q co-deletion status prediction, \gls{ece} and \gls{nll} both increased after dropout 0.20, but \gls{ece} showed smaller local fluctuations at higher dropout rates while \gls{nll} and \gls{auc} indicated clearer degradation. For tumor grade prediction, the difference between the metrics was most apparent: \gls{ece} showed local decreases at intermediate dropout values, particularly around dropout rates 0.25 and 0.45, whereas \gls{nll} increased more steadily and \gls{auc} decreased at higher dropout rates.

\begin{figure}[h!]
    \centering
    \includegraphics[width=1\linewidth]{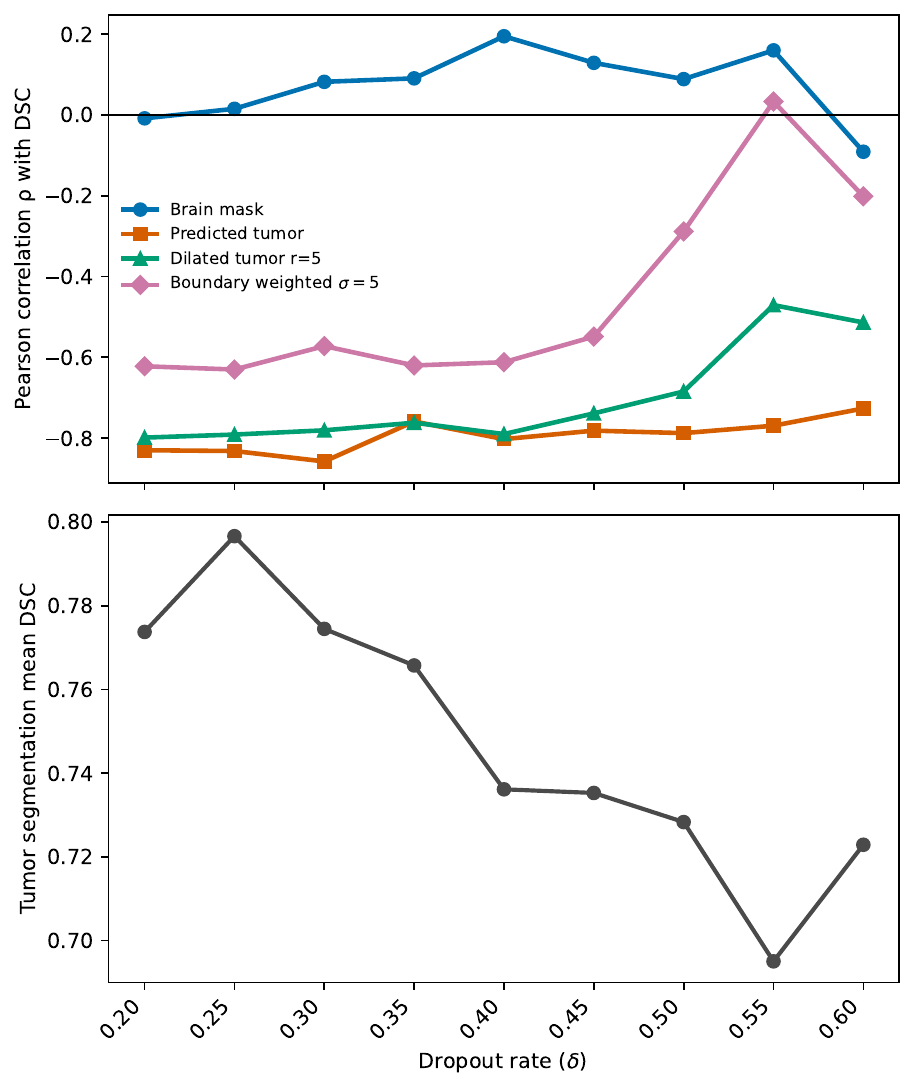}
     \caption{Tumor segmentation uncertainty quality analysis. The top panel presents the Pearson correlation coefficient between case-level predictive uncertainty and Dice Similarity Coefficient (DSC) as a function of the dropout rate $\delta$ used for the \gls{mcd} ensemble. Correlations are shown for four spatial aggregation strategies: brain-mask aggregation, predicted-tumor aggregation, 5-voxel dilated predicted-tumor aggregation, and boundary-weighted aggregation using $\sigma=5$ voxels. More negative correlations indicate that higher case-level uncertainty is associated with lower segmentation accuracy. The bottom panel presents the mean DSC between the predicted tumor segmentation and the ground truth in the test set as a function of the dropout rate $\delta$.}    \label{fig:seg_calib} 

\end{figure}

Figure~\ref{fig:seg_calib} presents the uncertainty quality analysis for the tumor segmentation task. Brain mask-based aggregation showed weak correlations with \gls{dsc} across dropout rates, with values fluctuating around zero and no consistent negative association. In contrast, predicted tumor aggregation showed the strongest and most stable negative correlations across the evaluated dropout range. Dilated tumor aggregation followed a similar trend at low and moderate dropout rates, but the correlation became progressively weaker at higher dropout values. Boundary-weighted aggregation also showed a negative association with \gls{dsc} at low and moderate dropout rates, although this association weakened more markedly as dropout increased. The lower panel shows that the mean segmentation \gls{dsc} was highest at moderate dropout rates, peaking at $0.25$. It generally decreased with stronger dropout. Based on its consistently strong negative association with \gls{dsc} and its lack of additional aggregation hyperparameters, predicted tumor aggregation was selected for the downstream segmentation uncertainty analyses.

Our evaluation across tasks and metrics indicated that the \gls{mcd} model with a dropout rate of 0.25 provided a favorable trade-off between predictive performance and uncertainty calibration, and was therefore selected for the remaining evaluation.

 To further assess whether the reported Pearson correlations reflected the case-wise uncertainty and performance relationship, we visualized predictive uncertainty against tumor segmentation \gls{dsc} at the selected configuration ($T=30$, $\delta=0.25$). The corresponding scatter plots are provided in Appendix~\ref{app:segmentation-correlation}. For predicted tumor and dilated tumor aggregation, the scatter plots showed a clear decreasing trend, consistent with the strong negative Pearson and Spearman correlations. Boundary-weighted aggregation also showed a negative monotonic trend, although with a weaker linear association. In contrast, brain mask aggregation showed no clear relationship between predictive uncertainty and \gls{dsc}. Overall, Pearson and Spearman correlation values were similar across the aggregation strategies, indicating that the main findings were not dependent on the use of a strictly linear association measure.

\begin{figure}[h!]
    \centering
    \includegraphics[width=1\linewidth]{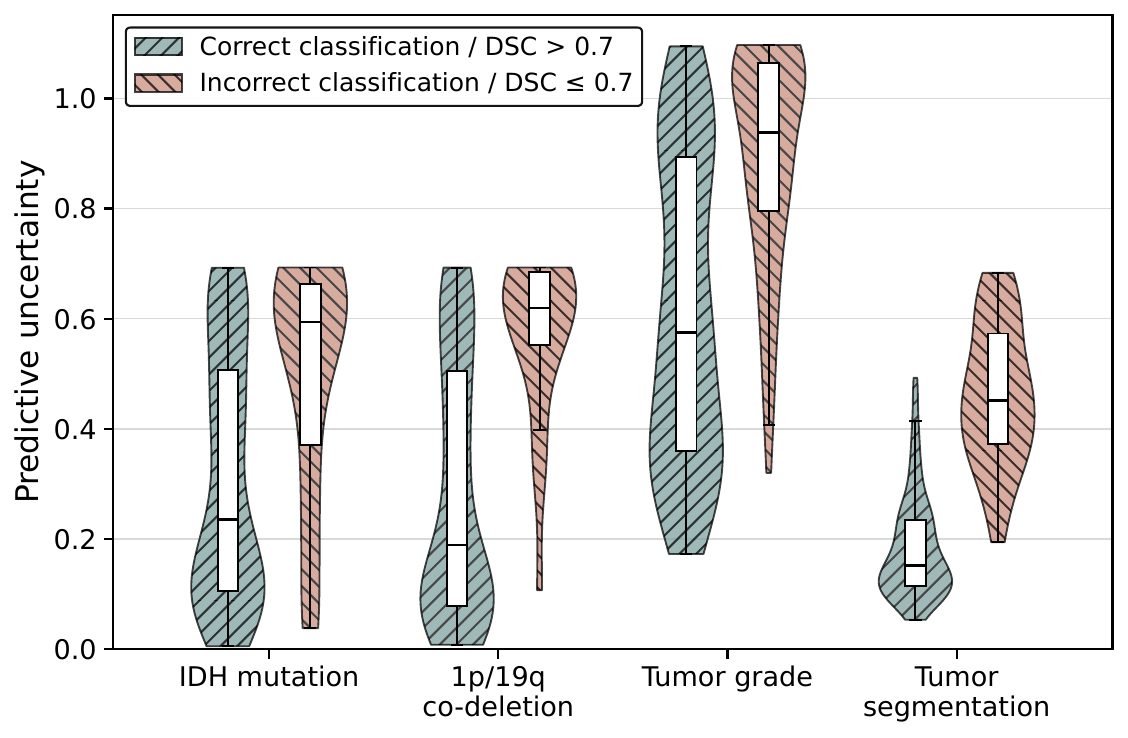}
    \caption{Violin plots of the predictive uncertainty across different tasks, comparing the distributions of high quality and low quality predictions, obtained with the optimal configuration of the \gls{mcd} model at $T=30$ and $\delta=0.25$. For the classification tasks, high and low quality predictions correspond to correctly and incorrectly classified cases, respectively. For the tumor segmentation task, high and low quality predictions represent cases with $ DSC >0.7$ and $DSC \leq 0.7$, respectively. For all tasks, the difference between the distributions for both groups was statistically significant (Mann--Whitney U test).}
    \label{fig:visual_calibration}
\end{figure}

Figure \ref{fig:visual_calibration} further highlights the ability of uncertainty scores to discriminate between high- and low-quality predictions, with low-quality predictions generally exhibiting higher uncertainty across tasks. The difference between the distributions for all tasks was statistically significant according to the Mann--Whitney U test.

In order to contextualize the uncertainty-based error detection analyses, we additionally reported the predictive performance of the selected \gls{mcd} configuration used throughout the remaining experiments ($T=30$, $\delta=0.25$). Table~\ref{tab:predictive-performance-selected-mcd} reports the task-specific class distribution in the test set, along with different performance metrics, including accuracy, balanced accuracy, macro-averaged F1-score and \gls{auc}. The class distributions highlight substantial imbalance, particularly for 1p/19q co-deletion status and tumor grade prediction tasks.

\begin{table}[h]
\centering

\caption{Predictive performance at the selected \gls{mcd} configuration ($T=30$, $\delta=0.25$) for the three classification tasks. For each task, the class distribution is also reported. Values are shown with $95\%$ confidence intervals obtained by $1000\times$ bootstrap resampling of the test set. \gls{auc} is reported at task level: for tumor grade, it corresponds to the macro-averaged one-vs-rest \gls{auc} across grade classes. Total test set size differs across tasks because molecular and histopathological annotations were not available for all patients. Arrows in metric column headers indicate that higher values are preferable ($\uparrow$).
}
\label{tab:predictive-performance-selected-mcd}
\scriptsize
\setlength{\tabcolsep}{1.5pt}
\resizebox{\columnwidth}{!}{
\begin{tabular}{@{}llcccc@{}}
\toprule
Task 
& \begin{tabular}[c]{@{}c@{}}Class\\distribution\end{tabular}
& Accuracy $\uparrow$ 
& \begin{tabular}[c]{@{}c@{}}Balanced\\accuracy $\uparrow$\end{tabular}
& \begin{tabular}[c]{@{}c@{}}Macro F1\\score $\uparrow$\end{tabular}
& \begin{tabular}[c]{@{}c@{}}AUC $\uparrow$\end{tabular} \\ 
\midrule

IDH 
& \begin{tabular}[c]{@{}l@{}}Wildtype: 129\\Mutated: 85\end{tabular}
& \begin{tabular}[c]{@{}c@{}}0.82\\{[0.77,0.87]}\end{tabular}
& \begin{tabular}[c]{@{}c@{}}0.79\\{[0.73,0.84]}\end{tabular}
& \begin{tabular}[c]{@{}c@{}}0.80\\{[0.74,0.85]}\end{tabular}
& \begin{tabular}[c]{@{}c@{}}0.90\\{[0.85,0.94]}\end{tabular}
\\
\addlinespace[3pt]

1p/19q
& \begin{tabular}[c]{@{}l@{}}Intact: 205\\Co-deleted: 25\end{tabular}
& \begin{tabular}[c]{@{}c@{}}0.90\\{[0.85,0.93]}\end{tabular}
& \begin{tabular}[c]{@{}c@{}}0.56\\{[0.50,0.63]}\end{tabular}
& \begin{tabular}[c]{@{}c@{}}0.57\\{[0.47,0.67]}\end{tabular}
& \begin{tabular}[c]{@{}c@{}}0.85\\{[0.77,0.91]}\end{tabular}
\\
\addlinespace[3pt]

Grade
& \begin{tabular}[c]{@{}l@{}}Grade 2: 45\\Grade 3: 58\\Grade 4: 132\end{tabular}
& \begin{tabular}[c]{@{}c@{}}0.71\\{[0.64,0.77]}\end{tabular}
& \begin{tabular}[c]{@{}c@{}}0.60\\{[0.57,0.64]}\end{tabular}
& \begin{tabular}[c]{@{}c@{}}0.51\\{[0.47,0.54]}\end{tabular}
& \begin{tabular}[c]{@{}c@{}}0.85\\{[0.81,0.88]}\end{tabular}
\\

\bottomrule
\end{tabular}
}

\end{table}

\subsection{Operational Utility of Uncertainty Estimates}

For tumor segmentation, the primary error definition used $\tau_{\mathrm{DSC}}=0.70$. A sensitivity analysis using different values of $\tau_{\mathrm{DSC}}\in\{0.60,0.70,0.80\}$ is reported in Appendix \ref{app:segmentation-sensitivity-analysis}.

Table~\ref{tab:uncertainty_quality} summarizes the ability of predictive, aleatoric, and epistemic uncertainty estimates to identify erroneous predictions across all tasks, evaluated using \gls{u-auc}, \gls{ap}, Lift, and \gls{aurc}. For each metric, $1000\times$ bootstrap confidence intervals are reported in the table to indicate the variability of the estimates. The observed error rates of 0.18, 0.10, 0.29, and 0.17 for \gls{idh} mutation status, 1p/19q co-deletion status, tumor grade and tumor segmentation, respectively, correspond to the baseline \gls{ap} for a random classifier.

For \gls{idh} mutation status prediction, predictive uncertainty achieved an \gls{ap} of 0.42, corresponding to a Lift of 2.37 relative to the baseline error rate. Aleatoric and epistemic uncertainty yielded lower \gls{ap} values of 0.32 and 0.34, corresponding to Lifts of 1.79 and 1.89, respectively. \gls{aurc} values were similar across uncertainty components, with all three components achieving an \gls{aurc} of 0.10. Despite aleatoric uncertainty being dominant in magnitude (as shown in Figure~\ref{fig:effect_mcd}), predictive uncertainty provided the numerically highest Lift.

For 1p/19q co-deletion status prediction, predictive and aleatoric uncertainty achieved \gls{ap} values of 0.37 and 0.39, corresponding to Lifts of 3.55 and 3.70, respectively, while epistemic uncertainty yielded a lower \gls{ap} of 0.16 (Lift 1.54). Predictive and aleatoric uncertainty also achieved lower \gls{aurc} values of 0.03, compared with 0.05 for epistemic uncertainty. In this task, the dominant aleatoric component also achieved the numerically highest Lift.

For tumor grade prediction, predictive and aleatoric uncertainty achieved identical \gls{ap} values of 0.59, corresponding to Lifts of 2.00, whereas epistemic uncertainty achieved a lower \gls{ap} of 0.45 (Lift 1.52). Similarly, predictive and aleatoric uncertainty achieved lower \gls{aurc} values of 0.12, compared with 0.17 for epistemic uncertainty.

For tumor segmentation, aleatoric uncertainty achieved the highest \gls{ap} of 0.83 (Lift 5.01), followed closely by predictive uncertainty with an \gls{ap} of 0.82 (Lift 4.95) and epistemic uncertainty with an \gls{ap} of 0.81 (Lift 4.90). All three uncertainty components achieved high \gls{u-auc} values, ranging from 0.95 to 0.96, and similar \gls{aurc} values of 0.13. Although epistemic uncertainty was dominant in magnitude (as shown in Figure~\ref{fig:effect_mcd}), aleatoric and predictive uncertainty still achieved comparable or higher Lift.

Overall, while all uncertainty types identified errors better than expected from the baseline error rates, the component with the highest magnitude (as shown in Figure~\ref{fig:effect_mcd}) did not consistently correspond to the highest Lift  or lowest \gls{aurc}. For \gls{idh} mutation status prediction, predictive uncertainty had the highest point estimates for AP and Lift, but the corresponding confidence intervals overlapped with those of aleatoric and epistemic uncertainty. For 1p/19q co-deletion status prediction and tumor grade prediction, predictive and aleatoric uncertainty showed very similar performance, whereas epistemic uncertainty generally showed lower AP, lower Lift, and higher \gls{aurc}. For tumor segmentation, all three uncertainty components showed comparable performance, with \gls{u-auc} values between 0.95 and 0.96, AP values between 0.81 and 0.83, Lift values between 4.90 and 5.01, and identical \gls{aurc} values of 0.13. Therefore, it can be noted that predictive uncertainty achieved performance comparable to that of individual decomposed components across all tasks given the high overlap between the confidence intervals. 

\begin{table*}[t]
\centering
\begingroup

\scriptsize
\setlength{\tabcolsep}{1.5pt}
\caption{Operational utility of uncertainty estimates per task. Assessment is made using Uncertainty-based Receiver Operating Characteristic Area Under the Curve (U-AUC), Average Precision (AP), Lift, and Area Under the Risk--Coverage Curve (AURC), derived from using uncertainty as a predictor of errors. Values are reported with $95\%$ confidence intervals obtained by $1000\times$ bootstrap resampling of the test set. The numerically highest values for \gls{u-auc}, AP, and Lift, and the numerically lowest values for \gls{aurc}, are indicated in bold for each task. Arrows in metric column headers indicate whether higher ($\uparrow$) or lower ($\downarrow$) values are preferable.}
\label{tab:uncertainty_quality}

\resizebox{\textwidth}{!}{
\begin{tabular}{l|c|cccc|cccc|cccc}
\hline
\multicolumn{1}{c|}{\multirow{3}{*}{Task}}
& \multirow{3}{*}{\begin{tabular}[c]{@{}c@{}}Error\\rate $\downarrow$\end{tabular}}
& \multicolumn{12}{c}{Uncertainty} \\

\cline{3-14}
&
& \multicolumn{4}{c|}{Predictive}
& \multicolumn{4}{c|}{Aleatoric}
& \multicolumn{4}{c}{Epistemic} \\

\cline{3-14}
&
& \begin{tabular}[c]{@{}c@{}}U-AUC $\uparrow$\end{tabular}
& AP $\uparrow$
& Lift $\uparrow$
& \gls{aurc} $\downarrow$
& \begin{tabular}[c]{@{}c@{}}U-AUC $\uparrow$\end{tabular}
& AP $\uparrow$
& Lift $\uparrow$
& \gls{aurc} $\downarrow$
& \begin{tabular}[c]{@{}c@{}}U-AUC $\uparrow$\end{tabular}
& AP $\uparrow$
& Lift $\uparrow$
& \gls{aurc} $\downarrow$
\\ \hline

\begin{tabular}[c]{@{}l@{}}\gls{idh} mutation\\status prediction\end{tabular}
& 0.18
& \begin{tabular}[c]{@{}c@{}}\textbf{0.73}\\{[0.62,0.82]}\end{tabular}
& \begin{tabular}[c]{@{}c@{}}\textbf{0.42}\\{[0.28,0.57]}\end{tabular}
& \begin{tabular}[c]{@{}c@{}}\textbf{2.37}\\{[1.72,3.34]}\end{tabular}
& \begin{tabular}[c]{@{}c@{}}\textbf{0.10}\\{[0.05,0.15]}\end{tabular}
& \begin{tabular}[c]{@{}c@{}}0.71\\{[0.61,0.80]}\end{tabular}
& \begin{tabular}[c]{@{}c@{}}0.32\\{[0.22,0.46]}\end{tabular}
& \begin{tabular}[c]{@{}c@{}}1.79\\{[1.39,2.54]}\end{tabular}
& \begin{tabular}[c]{@{}c@{}}\textbf{0.10}\\{[0.05,0.15]}\end{tabular}
& \begin{tabular}[c]{@{}c@{}}0.71\\{[0.61,0.80]}\end{tabular}
& \begin{tabular}[c]{@{}c@{}}0.34\\{[0.23,0.50]}\end{tabular}
& \begin{tabular}[c]{@{}c@{}}1.89\\{[1.45,2.80]}\end{tabular}
& \begin{tabular}[c]{@{}c@{}}\textbf{0.10}\\{[0.05,0.16]}\end{tabular}
\\

\begin{tabular}[c]{@{}l@{}}1p/19q co-deletion\\status prediction\end{tabular}
& 0.10
& \begin{tabular}[c]{@{}c@{}}\textbf{0.84}\\{[0.76,0.91]}\end{tabular}
& \begin{tabular}[c]{@{}c@{}}0.37\\{[0.22,0.58]}\end{tabular}
& \begin{tabular}[c]{@{}c@{}}3.55\\{[2.37,6.07]}\end{tabular}
& \begin{tabular}[c]{@{}c@{}}\textbf{0.03}\\{[0.01,0.05]}\end{tabular}
& \begin{tabular}[c]{@{}c@{}}\textbf{0.84}\\{[0.77,0.91]}\end{tabular}
& \begin{tabular}[c]{@{}c@{}}\textbf{0.39}\\{[0.22,0.58]}\end{tabular}
& \begin{tabular}[c]{@{}c@{}}\textbf{3.70}\\{[2.51,6.19]}\end{tabular}
& \begin{tabular}[c]{@{}c@{}}\textbf{0.03}\\{[0.01,0.04]}\end{tabular}
& \begin{tabular}[c]{@{}c@{}}0.68\\{[0.58,0.77]}\end{tabular}
& \begin{tabular}[c]{@{}c@{}}0.16\\{[0.10,0.29]}\end{tabular}
& \begin{tabular}[c]{@{}c@{}}1.54\\{[1.22,2.59]}\end{tabular}
& \begin{tabular}[c]{@{}c@{}}0.05\\{[0.03,0.08]}\end{tabular}
\\

\begin{tabular}[c]{@{}l@{}}Tumor grade\\prediction\end{tabular}
& 0.29
& \begin{tabular}[c]{@{}c@{}}0.78\\{[0.72,0.84]}\end{tabular}
& \begin{tabular}[c]{@{}c@{}}\textbf{0.59}\\{[0.47,0.71]}\end{tabular}
& \begin{tabular}[c]{@{}c@{}}\textbf{2.00}\\{[1.66,2.47]}\end{tabular}
& \begin{tabular}[c]{@{}c@{}}\textbf{0.12}\\{[0.09,0.17]}\end{tabular}
& \begin{tabular}[c]{@{}c@{}}\textbf{0.79}\\{[0.73,0.85]}\end{tabular}
& \begin{tabular}[c]{@{}c@{}}\textbf{0.59}\\{[0.47,0.72]}\end{tabular}
& \begin{tabular}[c]{@{}c@{}}\textbf{2.00}\\{[1.68,2.50]}\end{tabular}
& \begin{tabular}[c]{@{}c@{}}\textbf{0.12}\\{[0.09,0.17]}\end{tabular}
& \begin{tabular}[c]{@{}c@{}}0.70\\{[0.62,0.77]}\end{tabular}
& \begin{tabular}[c]{@{}c@{}}0.45\\{[0.35,0.58]}\end{tabular}
& \begin{tabular}[c]{@{}c@{}}1.52\\{[1.28,1.94]}\end{tabular}
& \begin{tabular}[c]{@{}c@{}}0.17\\{[0.12,0.24]}\end{tabular}
\\

\begin{tabular}[c]{@{}l@{}}Tumor\\segmentation\end{tabular}
& 0.17
& \begin{tabular}[c]{@{}c@{}}0.95\\{[0.92,0.98]}\end{tabular}
& \begin{tabular}[c]{@{}c@{}}0.82\\{[0.72,0.91]}\end{tabular}
& \begin{tabular}[c]{@{}c@{}}4.95\\{[3.89,6.67]}\end{tabular}
& \begin{tabular}[c]{@{}c@{}}\textbf{0.13}\\{[0.12,0.14]}\end{tabular}
& \begin{tabular}[c]{@{}c@{}}\textbf{0.96}\\{[0.94,0.98]}\end{tabular}
& \begin{tabular}[c]{@{}c@{}}\textbf{0.83}\\{[0.72,0.91]}\end{tabular}
& \begin{tabular}[c]{@{}c@{}}\textbf{5.01}\\{[3.93,6.80]}\end{tabular}
& \begin{tabular}[c]{@{}c@{}}\textbf{0.13}\\{[0.12,0.14]}\end{tabular}
& \begin{tabular}[c]{@{}c@{}}0.95\\{[0.92,0.98]}\end{tabular}
& \begin{tabular}[c]{@{}c@{}}0.81\\{[0.71,0.90]}\end{tabular}
& \begin{tabular}[c]{@{}c@{}}4.90\\{[3.85,6.64]}\end{tabular}
& \begin{tabular}[c]{@{}c@{}}\textbf{0.13}\\{[0.12,0.14]}\end{tabular}
\\ \hline

\end{tabular}
}
\endgroup
\end{table*}

\subsection{Comparison of MCD with other UQ methods}
\label{sec:mcd_other_uq_methods}

After selecting the final \gls{mcd} configuration, we compared \gls{mcd} with two ensemble-based \gls{uq} methods, \gls{de} and \gls{mcde}, using the same operational-utility framework. Table~\ref{tab:uq_method_predictive_comparison} reports the performance of predictive uncertainty for error detection and selective prediction across the three classification tasks and tumor segmentation using predicted-tumor aggregation.

For the classification tasks, all three methods produced uncertainty estimates with Lift values above 1, indicating better-than-baseline error identification. For \gls{idh} mutation status prediction, \gls{mcde} had the lowest \gls{aurc}, while \gls{mcd} achieved the highest AP and Lift. For 1p/19q co-deletion status prediction, \gls{de} and \gls{mcd} showed the same \gls{u-auc} values, with \gls{de} achieving the highest AP and Lift. For tumor grade prediction, \gls{de} achieved the highest \gls{u-auc}, AP, and Lift, whereas \gls{mcde} achieved the lowest error rate and \gls{aurc}.

For tumor segmentation, predictive uncertainty was aggregated within the predicted tumor region. All three methods achieved high \gls{u-auc} values, indicating that predicted-tumor uncertainty was informative for identifying low-quality segmentations. \gls{mcde} achieved the highest \gls{u-auc} and Lift, whereas \gls{mcd} achieved the highest AP. The lowest \gls{aurc} values were observed for \gls{de} and \gls{mcde}.

Overall, ensemble-based uncertainty estimates provided operational utility comparable to \gls{mcd}, but no method consistently dominated across all tasks and metrics. In particular, \gls{mcde} tended to provide favorable \gls{aurc} values, suggesting slightly improved selective-risk behavior, whereas \gls{de} or \gls{mcd} more often achieved higher AP or Lift. Because confidence intervals overlapped in several comparisons, these differences were interpreted as task- and metric-dependent trends rather than as evidence of consistent superiority of one uncertainty method. Therefore, the subsequent analyses were based on \gls{mcd}.

\begin{table}[t]
\centering
\begingroup

\scriptsize
\setlength{\tabcolsep}{1.5pt}
\caption{Comparison of uncertainty quantification methods using predictive uncertainty. Assessment is made using Uncertainty-based Receiver Operating Characteristic Area Under the Curve (U-AUC), Average Precision (AP), Lift, and Area Under the Risk--Coverage Curve (AURC), derived from using predictive uncertainty as a predictor of errors. Values are reported with $95\%$ confidence intervals obtained by $1000\times$ bootstrap resampling of the test set. For tumor segmentation, predictive uncertainty was aggregated within the predicted tumor region. The numerically highest values for \gls{u-auc}, AP, and Lift, and the numerically lowest values for \gls{aurc}, are indicated in bold for each task. Arrows in metric column headers indicate whether higher ($\uparrow$) or lower ($\downarrow$) values are preferable.}
\label{tab:uq_method_predictive_comparison}

\resizebox{\columnwidth}{!}{
\begin{tabular}{llccccc}
\hline
Task & Method & Error $\downarrow$ & U-AUC $\uparrow$ & AP $\uparrow$ & Lift $\uparrow$ & \gls{aurc} $\downarrow$ \\
\hline

\multirow{3}{*}{\begin{tabular}[c]{@{}l@{}}\gls{idh}\\mutation\end{tabular}}
& \gls{de}
& \textbf{0.17}
& \begin{tabular}[c]{@{}c@{}}0.72\\{[0.62,0.82]}\end{tabular}
& \begin{tabular}[c]{@{}c@{}}0.37\\{[0.24,0.56]}\end{tabular}
& \begin{tabular}[c]{@{}c@{}}2.15\\{[1.57,3.23]}\end{tabular}
& \begin{tabular}[c]{@{}c@{}}0.10\\{[0.05,0.15]}\end{tabular}
\\

& \gls{mcd}
& 0.18
& \begin{tabular}[c]{@{}c@{}}0.73\\{[0.62,0.82]}\end{tabular}
& \begin{tabular}[c]{@{}c@{}}\textbf{0.42}\\{[0.28,0.57]}\end{tabular}
& \begin{tabular}[c]{@{}c@{}}\textbf{2.37}\\{[1.72,3.34]}\end{tabular}
& \begin{tabular}[c]{@{}c@{}}0.10\\{[0.05,0.15]}\end{tabular}
\\

& \gls{mcde}
& \textbf{0.17}
& \begin{tabular}[c]{@{}c@{}}\textbf{0.73}\\{[0.63,0.83]}\end{tabular}
& \begin{tabular}[c]{@{}c@{}}0.34\\{[0.23,0.50]}\end{tabular}
& \begin{tabular}[c]{@{}c@{}}2.03\\{[1.53,2.98]}\end{tabular}
& \begin{tabular}[c]{@{}c@{}}\textbf{0.09}\\{[0.05,0.14]}\end{tabular}
\\
\hline

\multirow{3}{*}{\begin{tabular}[c]{@{}l@{}}1p/19q\\co-deletion\end{tabular}}
& \gls{de}
& 0.10
& \begin{tabular}[c]{@{}c@{}}\textbf{0.84}\\{[0.76,0.91]}\end{tabular}
& \begin{tabular}[c]{@{}c@{}}\textbf{0.39}\\{[0.22,0.59]}\end{tabular}
& \begin{tabular}[c]{@{}c@{}}\textbf{3.90}\\{[2.53,6.62]}\end{tabular}
& \begin{tabular}[c]{@{}c@{}}\textbf{0.03}\\{[0.01,0.04]}\end{tabular}
\\

& \gls{mcd}
& 0.10
& \begin{tabular}[c]{@{}c@{}}\textbf{0.84}\\{[0.76,0.91]}\end{tabular}
& \begin{tabular}[c]{@{}c@{}}0.37\\{[0.22,0.58]}\end{tabular}
& \begin{tabular}[c]{@{}c@{}}3.55\\{[2.37,6.07]}\end{tabular}
& \begin{tabular}[c]{@{}c@{}}\textbf{0.03}\\{[0.01,0.05]}\end{tabular}
\\

& \gls{mcde}
& \textbf{0.09}
& \begin{tabular}[c]{@{}c@{}}0.81\\{[0.74,0.89]}\end{tabular}
& \begin{tabular}[c]{@{}c@{}}0.26\\{[0.16,0.45]}\end{tabular}
& \begin{tabular}[c]{@{}c@{}}2.84\\{[2.06,4.91]}\end{tabular}
& \begin{tabular}[c]{@{}c@{}}\textbf{0.03}\\{[0.01,0.04]}\end{tabular}
\\
\hline

\multirow{3}{*}{\begin{tabular}[c]{@{}l@{}}Tumor\\grade\end{tabular}}
& \gls{de}
& 0.30
& \begin{tabular}[c]{@{}c@{}}\textbf{0.80}\\{[0.73,0.86]}\end{tabular}
& \begin{tabular}[c]{@{}c@{}}\textbf{0.61}\\{[0.50,0.73]}\end{tabular}
& \begin{tabular}[c]{@{}c@{}}\textbf{2.04}\\{[1.70,2.53]}\end{tabular}
& \begin{tabular}[c]{@{}c@{}}0.13\\{[0.09,0.18]}\end{tabular}
\\

& \gls{mcd}
& 0.29
& \begin{tabular}[c]{@{}c@{}}0.78\\{[0.72,0.84]}\end{tabular}
& \begin{tabular}[c]{@{}c@{}}0.59\\{[0.47,0.71]}\end{tabular}
& \begin{tabular}[c]{@{}c@{}}2.00\\{[1.66,2.47]}\end{tabular}
& \begin{tabular}[c]{@{}c@{}}0.12\\{[0.09,0.17]}\end{tabular}
\\

& \gls{mcde}
& \textbf{0.28}
& \begin{tabular}[c]{@{}c@{}}0.78\\{[0.72,0.84]}\end{tabular}
& \begin{tabular}[c]{@{}c@{}}0.51\\{[0.40,0.65]}\end{tabular}
& \begin{tabular}[c]{@{}c@{}}1.82\\{[1.53,2.30]}\end{tabular}
& \begin{tabular}[c]{@{}c@{}}\textbf{0.12}\\{[0.08,0.17]}\end{tabular}
\\
\hline

\multirow{3}{*}{\begin{tabular}[c]{@{}l@{}}Tumor\\segmentation\end{tabular}}
& \gls{de}
& 0.13
& \begin{tabular}[c]{@{}c@{}}0.94\\{[0.90,0.96]}\end{tabular}
& \begin{tabular}[c]{@{}c@{}}0.65\\{[0.48,0.82]}\end{tabular}
& \begin{tabular}[c]{@{}c@{}}4.91\\{[3.70,7.30]}\end{tabular}
& \begin{tabular}[c]{@{}c@{}}\textbf{0.12}\\{[0.11,0.13]}\end{tabular}
\\

& \gls{mcd}
& 0.17
& \begin{tabular}[c]{@{}c@{}}0.95\\{[0.92,0.98]}\end{tabular}
& \begin{tabular}[c]{@{}c@{}}\textbf{0.82}\\{[0.72,0.91]}\end{tabular}
& \begin{tabular}[c]{@{}c@{}}4.95\\{[3.89,6.67]}\end{tabular}
& \begin{tabular}[c]{@{}c@{}}0.13\\{[0.12,0.14]}\end{tabular}
\\

& \gls{mcde}
& \textbf{0.12}
& \begin{tabular}[c]{@{}c@{}}\textbf{0.96}\\{[0.92,0.98]}\end{tabular}
& \begin{tabular}[c]{@{}c@{}}0.79\\{[0.65,0.89]}\end{tabular}
& \begin{tabular}[c]{@{}c@{}}\textbf{6.37}\\{[4.82,9.16]}\end{tabular}
& \begin{tabular}[c]{@{}c@{}}\textbf{0.12}\\{[0.12,0.13]}\end{tabular}
\\
\hline

\end{tabular}
}
\endgroup
\end{table}

The corresponding results for aleatoric and epistemic uncertainty are reported in Appendix ~\ref{app:comparison-uq}, Table ~\ref{tab:uq_method_components_comparison_appendix}. For aleatoric uncertainty, the trends were broadly consistent with the predictive-uncertainty analysis: all three methods showed operational utility above the baseline error rate across the classification tasks, and tumor segmentation uncertainty aggregated within the predicted tumor region remained highly informative.
For epistemic uncertainty, classification error-detection performance was weaker and less consistent than for predictive or aleatoric uncertainty. This was most evident for 1p/19q co-deletion status prediction, where epistemic uncertainty showed lower \gls{u-auc}, AP, and Lift than the corresponding predictive and aleatoric uncertainty estimates. Across the classification tasks, \gls{mcd}-based epistemic uncertainty tended to retain the most consistent operational utility, whereas the ensemble-based epistemic estimates showed more metric-dependent behavior.

\subsection{Interaction Between Tumor Segmentation and Classification}

\begin{figure}[t]
\centering
\includegraphics[width=1\linewidth]{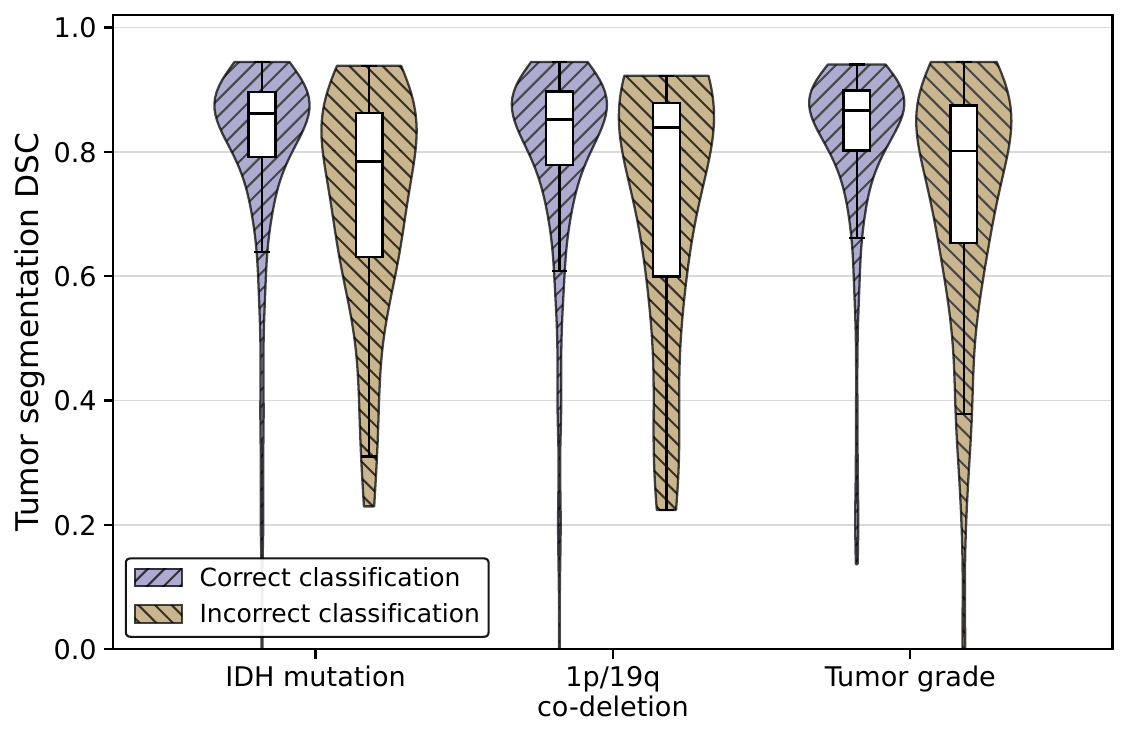}
\caption{Distribution of tumor segmentation Dice Similarity Coefficient (DSC) scores for correctly and incorrectly classified cases for the IDH mutation status, 1p/19q co-deletion status and tumor grade. Cases are stratified according to whether the corresponding prediction was correct or incorrect. For each task, higher DSC values indicate better agreement between predicted and reference tumor segmentations. Statistical significance between groups was found for the IDH mutation status and tumor grade prediction (Mann--Whitney U test).}
\label{fig:violin_dice_classification}
\end{figure}

Figure~\ref{fig:violin_dice_classification} shows the distribution of tumor segmentation \gls{dsc} scores stratified by classification correctness for each classification task. For \gls{idh} mutation status, correctly classified cases exhibited higher segmentation performance compared to misclassified cases, and this difference was statistically significant (Mann–Whitney U test). For 1p/19q co-deletion status, correctly classified cases tended to present slightly higher \gls{dsc} scores; however, the difference did not reach statistical significance. For tumor grade, correctly classified cases again showed higher \gls{dsc} scores, with a statistically significant difference between groups (Mann–Whitney U test).


Table~\ref{tab:uncertainty_correlation} presents the correlations between tumor segmentation uncertainty and classification uncertainty. Overall, predictive and aleatoric uncertainty components exhibited positive correlations across the classification tasks, indicating that cases with higher tumor segmentation uncertainty also tended to present higher classification uncertainty. These positive associations were supported by confidence intervals that remained above zero for the predictive and aleatoric components across all three classification tasks.

\begin{table}[h]
\centering
\begingroup

\caption{Pearson correlation ($\rho$) between tumor segmentation uncertainty and classification uncertainty for each task. Correlations were computed separately for predictive, aleatoric, and epistemic uncertainty components. Values are reported with $95\%$ confidence intervals obtained by $1000\times$ bootstrap resampling of the test set. The numerically highest correlation coefficient per task is indicated in bold.}
\label{tab:uncertainty_correlation}

\scriptsize
\setlength{\tabcolsep}{4pt}
\begin{tabular}{lccc}
\toprule
\multicolumn{1}{c}{\multirow{2}{*}{Task}}
& \multicolumn{3}{c}{Uncertainty} \\
\cmidrule(lr){2-4}
& Predictive & Aleatoric & Epistemic \\
\midrule

\begin{tabular}[c]{@{}l@{}}\gls{idh} mutation\\status prediction\end{tabular}
& \begin{tabular}[c]{@{}c@{}}$0.48$\\{[0.38, 0.57]}\end{tabular}
& \begin{tabular}[c]{@{}c@{}}$\mathbf{0.51}$\\{[0.42, 0.58]}\end{tabular}
& \begin{tabular}[c]{@{}c@{}}$0.09$\\{[-0.04, 0.21]}\end{tabular}
\\

\begin{tabular}[c]{@{}l@{}}1p/19q co-deletion\\status prediction\end{tabular}
& \begin{tabular}[c]{@{}c@{}}$0.22$\\{[0.12, 0.33]}\end{tabular}
& \begin{tabular}[c]{@{}c@{}}$\mathbf{0.24}$\\{[0.15, 0.33]}\end{tabular}
& \begin{tabular}[c]{@{}c@{}}$-0.15$\\{[-0.25, -0.05]}\end{tabular}
\\

\begin{tabular}[c]{@{}l@{}}Tumor grade\\prediction\end{tabular}
& \begin{tabular}[c]{@{}c@{}}$\mathbf{0.53}$\\{[0.45, 0.61]}\end{tabular}
& \begin{tabular}[c]{@{}c@{}}$0.50$\\{[0.42, 0.57]}\end{tabular}
& \begin{tabular}[c]{@{}c@{}}$0.10$\\{[-0.03, 0.26]}\end{tabular}
\\

\bottomrule
\end{tabular}
\endgroup
\end{table}

For \gls{idh} mutation status prediction, tumor segmentation uncertainty showed moderate correlations with predictive and aleatoric classification uncertainty, with correlation coefficients of 0.48 and 0.51, respectively. The association with epistemic uncertainty was weaker, with a correlation coefficient of 0.09, and its confidence interval included zero, indicating no clear positive association for this component.

For 1p/19q co-deletion status prediction, correlations were weaker overall. Predictive and aleatoric uncertainty showed modest positive associations, with correlation coefficients of 0.22 and 0.24, respectively. 
In contrast, epistemic uncertainty showed a weak negative correlation of $-0.15$, and its confidence interval remained below zero. This indicates that, for 1p/19q co-deletion status prediction, higher tumor segmentation epistemic uncertainty was not associated with higher classification epistemic uncertainty.

For tumor grade prediction, tumor segmentation uncertainty was again moderately correlated with predictive and aleatoric classification uncertainty, with correlation coefficients of 0.53 and 0.50, respectively. The association with epistemic uncertainty remained weak, with a correlation coefficient of 0.10, and its confidence interval included zero. Overall, predictive and aleatoric uncertainty demonstrated consistent positive cross-task associations, whereas epistemic uncertainty showed weaker and less consistent relationships across tasks.


Table~\ref{tab:seg_uncertainty_predict_cls} summarizes the ability of tumor segmentation uncertainty to identify classification errors. Overall, tumor segmentation uncertainty provided moderate discrimination of classification errors, with \gls{u-auc} values ranging from 0.61 to 0.69. In terms of \gls{ap}, predictive segmentation uncertainty achieved values of 0.30, 0.23, and 0.44 for \gls{idh}, 1p/19q, and tumor grade prediction, corresponding to Lifts of 1.68, 2.16, and 1.52, respectively. Similar Lift values were observed for aleatoric and epistemic segmentation uncertainty, ranging from 1.45 to 2.55 depending on the task and component. For all tasks and uncertainty components, the lower bounds of the Lift confidence intervals remained above 1, indicating that high segmentation uncertainty identified subsets of cases with higher classification error rates than expected under random selection. \gls{aurc} values ranged from 0.06 to 0.22, with lower values indicating lower residual risk among retained cases.

Across tasks, no single segmentation uncertainty component was consistently best across all metrics. Aleatoric segmentation uncertainty achieved the highest \gls{u-auc} in all three classification tasks. However, the highest AP and Lift were obtained by epistemic uncertainty for \gls{idh} mutation status prediction, aleatoric uncertainty for 1p/19q co-deletion status prediction, and epistemic uncertainty for tumor grade prediction. In each task, the confidence intervals of the segmentation uncertainty components overlapped for \gls{u-auc}, AP, and Lift. 


Compared to classification uncertainty (Table~\ref{tab:uncertainty_quality}), tumor segmentation uncertainty showed lower error-detection and selective-prediction performance when comparing the predictive uncertainty scores directly. For \gls{idh} mutation status prediction, predictive segmentation uncertainty achieved a Lift of 1.68 and an \gls{aurc} of 0.15, whereas predictive classification uncertainty achieved a Lift of 2.37 and an \gls{aurc} of 0.10. For 1p/19q co-deletion status prediction, predictive segmentation uncertainty achieved a Lift of 2.16 and an \gls{aurc} of 0.06, compared with a Lift of 3.55 and an \gls{aurc} of 0.03 for predictive classification uncertainty. For tumor grade prediction, predictive segmentation uncertainty achieved a Lift of 1.52 and an \gls{aurc} of 0.22, compared with a Lift of 2.00 and an \gls{aurc} of 0.12 for predictive classification uncertainty. Thus, for all three classification tasks, predictive classification uncertainty achieved higher Lift and lower \gls{aurc} than predictive tumor segmentation uncertainty.

The same pattern was observed for aleatoric uncertainty, where classification uncertainty achieved higher Lift and lower \gls{aurc} than tumor segmentation uncertainty for all three classification tasks. For epistemic uncertainty, the comparison was less uniform for Lift: classification uncertainty achieved a higher Lift for \gls{idh} mutation status prediction, whereas tumor segmentation uncertainty had a higher Lift for 1p/19q co-deletion status and tumor grade prediction. However, classification epistemic uncertainty still achieved lower \gls{aurc} than tumor segmentation epistemic uncertainty for all three tasks. Overall, tumor segmentation uncertainty was informative for identifying classification errors, but classification-specific uncertainty remained the more direct and generally stronger error predictor.

The proposed trust score demonstrated consistent ability to identify classification errors across tasks (Table~\ref{tab:trust_score_performance}). For \gls{idh} mutation status prediction, the trust score achieved a \gls{u-auc} of 0.73 and an \gls{ap} of 0.38, corresponding to a Lift of 2.12 relative to the baseline error rate of 0.18, with an \gls{aurc} of 0.09. For 1p/19q co-deletion status prediction, the trust score yielded a \gls{u-auc} of 0.85 and an \gls{ap} of 0.42, representing a Lift of 3.98 over the baseline error rate of 0.10, with an \gls{aurc} of 0.02. For tumor grade prediction, the trust score achieved a \gls{u-auc} of 0.77 and an \gls{ap} of 0.51, corresponding to a Lift of 1.77 relative to the baseline error rate of 0.29, with an \gls{aurc} of 0.13. For all three tasks, the lower bounds of the Lift confidence intervals remained above 1, indicating that low-trust cases contained more classification errors than expected under random selection.

Compared with predictive classification uncertainty alone (Table~\ref{tab:uncertainty_quality}), the trust score did not show a consistent improvement across tasks. For \gls{idh} mutation status prediction, the trust score achieved the same \gls{u-auc} as predictive classification uncertainty, lower \gls{ap} and Lift, and slightly lower \gls{aurc}. For 1p/19q co-deletion status prediction, the trust score achieved higher point estimates for \gls{u-auc}, \gls{ap}, and Lift, and a lower \gls{aurc} than predictive classification uncertainty alone. In contrast, for tumor grade prediction, the trust score achieved lower \gls{u-auc}, \gls{ap}, and Lift, and higher \gls{aurc} than predictive classification uncertainty alone. The confidence intervals for the trust score overlapped with those of predictive classification uncertainty across these comparisons. Therefore, although the trust score slightly improved the point estimates for 1p/19q co-deletion status prediction, the results do not show a consistent improvement over predictive classification uncertainty alone across all classification tasks.

\begin{table*}[h]
\centering
\begingroup

\scriptsize
\setlength{\tabcolsep}{1.5pt}
\caption{Ability of tumor segmentation uncertainty to predict classification errors. Assessment is performed using Uncertainty-based Receiver Operating Characteristic Area Under the Curve (U-AUC), Average Precision (AP), Lift, and Area Under the Risk--Coverage Curve (AURC). Values are reported with $95\%$ confidence intervals obtained by $1000\times$ bootstrap resampling of the test set. The numerically highest values for \gls{u-auc}, AP, and Lift, and the numerically lowest values for \gls{aurc}, are indicated in bold for each task. Arrows in metric column headers indicate whether higher ($\uparrow$) or lower ($\downarrow$) values are preferable.}
\label{tab:seg_uncertainty_predict_cls}

\resizebox{\textwidth}{!}{
\begin{tabular}{l|c|cccc|cccc|cccc}
\hline
\multicolumn{1}{c|}{\multirow{3}{*}{Task}}
& \multirow{3}{*}{\begin{tabular}[c]{@{}c@{}}Error\\rate $\downarrow$\end{tabular}}
& \multicolumn{12}{c}{Tumor segmentation uncertainty as a predictor of classification errors} \\

\cline{3-14}
&
& \multicolumn{4}{c|}{Predictive}
& \multicolumn{4}{c|}{Aleatoric}
& \multicolumn{4}{c}{Epistemic} \\

\cline{3-14}
&
& \begin{tabular}[c]{@{}c@{}}U-AUC $\uparrow$\end{tabular}
& AP $\uparrow$
& Lift $\uparrow$
& \gls{aurc} $\downarrow$
& \begin{tabular}[c]{@{}c@{}}U-AUC $\uparrow$\end{tabular}
& AP $\uparrow$
& Lift $\uparrow$
& \gls{aurc} $\downarrow$
& \begin{tabular}[c]{@{}c@{}}U-AUC $\uparrow$\end{tabular}
& AP $\uparrow$
& Lift $\uparrow$
& \gls{aurc} $\downarrow$
\\ \hline

\begin{tabular}[c]{@{}l@{}}\gls{idh} mutation\\status prediction\end{tabular}
& 0.18
& \begin{tabular}[c]{@{}c@{}}0.62\\{[0.51,0.72]}\end{tabular}
& \begin{tabular}[c]{@{}c@{}}0.30\\{[0.21,0.46]}\end{tabular}
& \begin{tabular}[c]{@{}c@{}}1.68\\{[1.27,2.60]}\end{tabular}
& \begin{tabular}[c]{@{}c@{}}0.15\\{[0.08,0.22]}\end{tabular}
& \begin{tabular}[c]{@{}c@{}}\textbf{0.65}\\{[0.55,0.75]}\end{tabular}
& \begin{tabular}[c]{@{}c@{}}0.28\\{[0.20,0.40]}\end{tabular}
& \begin{tabular}[c]{@{}c@{}}1.54\\{[1.22,2.23]}\end{tabular}
& \begin{tabular}[c]{@{}c@{}}\textbf{0.13}\\{[0.07,0.20]}\end{tabular}
& \begin{tabular}[c]{@{}c@{}}0.61\\{[0.50,0.72]}\end{tabular}
& \begin{tabular}[c]{@{}c@{}}\textbf{0.32}\\{[0.22,0.46]}\end{tabular}
& \begin{tabular}[c]{@{}c@{}}\textbf{1.79}\\{[1.30,2.72]}\end{tabular}
& \begin{tabular}[c]{@{}c@{}}0.15\\{[0.09,0.23]}\end{tabular}
\\

\begin{tabular}[c]{@{}l@{}}1p/19q co-deletion\\status prediction\end{tabular}
& 0.10
& \begin{tabular}[c]{@{}c@{}}0.68\\{[0.57,0.78]}\end{tabular}
& \begin{tabular}[c]{@{}c@{}}0.23\\{[0.12,0.41]}\end{tabular}
& \begin{tabular}[c]{@{}c@{}}2.16\\{[1.38,4.15]}\end{tabular}
& \begin{tabular}[c]{@{}c@{}}\textbf{0.06}\\{[0.03,0.11]}\end{tabular}
& \begin{tabular}[c]{@{}c@{}}\textbf{0.69}\\{[0.57,0.79]}\end{tabular}
& \begin{tabular}[c]{@{}c@{}}\textbf{0.27}\\{[0.14,0.43]}\end{tabular}
& \begin{tabular}[c]{@{}c@{}}\textbf{2.55}\\{[1.50,4.56]}\end{tabular}
& \begin{tabular}[c]{@{}c@{}}\textbf{0.06}\\{[0.03,0.11]}\end{tabular}
& \begin{tabular}[c]{@{}c@{}}0.67\\{[0.56,0.78]}\end{tabular}
& \begin{tabular}[c]{@{}c@{}}0.24\\{[0.12,0.40]}\end{tabular}
& \begin{tabular}[c]{@{}c@{}}2.26\\{[1.36,4.19]}\end{tabular}
& \begin{tabular}[c]{@{}c@{}}\textbf{0.06}\\{[0.03,0.11]}\end{tabular}
\\

\begin{tabular}[c]{@{}l@{}}Tumor grade\\prediction\end{tabular}
& 0.29
& \begin{tabular}[c]{@{}c@{}}0.65\\{[0.57,0.73]}\end{tabular}
& \begin{tabular}[c]{@{}c@{}}0.44\\{[0.35,0.58]}\end{tabular}
& \begin{tabular}[c]{@{}c@{}}1.52\\{[1.25,1.94]}\end{tabular}
& \begin{tabular}[c]{@{}c@{}}0.22\\{[0.15,0.30]}\end{tabular}
& \begin{tabular}[c]{@{}c@{}}\textbf{0.67}\\{[0.59,0.74]}\end{tabular}
& \begin{tabular}[c]{@{}c@{}}0.42\\{[0.34,0.55]}\end{tabular}
& \begin{tabular}[c]{@{}c@{}}1.45\\{[1.23,1.82]}\end{tabular}
& \begin{tabular}[c]{@{}c@{}}\textbf{0.19}\\{[0.14,0.26]}\end{tabular}
& \begin{tabular}[c]{@{}c@{}}0.65\\{[0.56,0.73]}\end{tabular}
& \begin{tabular}[c]{@{}c@{}}\textbf{0.45}\\{[0.35,0.58]}\end{tabular}
& \begin{tabular}[c]{@{}c@{}}\textbf{1.54}\\{[1.27,1.95]}\end{tabular}
& \begin{tabular}[c]{@{}c@{}}0.22\\{[0.16,0.30]}\end{tabular}
\\ \hline

\end{tabular}
}
\endgroup
\end{table*}

\begin{table}[t]
\centering
\begingroup

\scriptsize
\setlength{\tabcolsep}{2pt}
\caption{Performance of the proposed trust score for predicting classification errors. Assessment is performed using Uncertainty-based Receiver Operating Characteristic Area Under the Curve (U-AUC), Average Precision (AP), Lift, and Area Under the Risk--Coverage Curve (AURC). Values are reported with $95\%$ confidence intervals obtained by $1000\times$ bootstrap resampling of the test set. Arrows in metric column headers indicate whether higher ($\uparrow$) or lower ($\downarrow$) values are preferable.}
\label{tab:trust_score_performance}

\begin{tabular}{lccccc}
\hline

\multicolumn{1}{c}{\multirow{2}{*}{Task}}
& \multirow{2}{*}{\begin{tabular}[c]{@{}c@{}}Error\\rate $\downarrow$\end{tabular}}
& \multicolumn{4}{c}{Trust score} \\

\cline{3-6}

&
& \begin{tabular}[c]{@{}c@{}}U-AUC $\uparrow$\end{tabular}
& AP $\uparrow$
& Lift $\uparrow$
& \gls{aurc} $\downarrow$

\\ \hline

\begin{tabular}[c]{@{}l@{}}\gls{idh} mutation\\status prediction\end{tabular}
& 0.18
& \begin{tabular}[c]{@{}c@{}}0.73\\{[0.64,0.81]}\end{tabular}
& \begin{tabular}[c]{@{}c@{}}0.38\\{[0.26,0.53]}\end{tabular}
& \begin{tabular}[c]{@{}c@{}}2.12\\{[1.58,3.10]}\end{tabular}
& \begin{tabular}[c]{@{}c@{}}0.09\\{[0.05,0.14]}\end{tabular}
\\

\begin{tabular}[c]{@{}l@{}}1p/19q co-deletion\\status prediction\end{tabular}
& 0.10
& \begin{tabular}[c]{@{}c@{}}0.85\\{[0.77,0.91]}\end{tabular}
& \begin{tabular}[c]{@{}c@{}}0.42\\{[0.25,0.58]}\end{tabular}
& \begin{tabular}[c]{@{}c@{}}3.98\\{[2.71,6.39]}\end{tabular}
& \begin{tabular}[c]{@{}c@{}}0.02\\{[0.01,0.04]}\end{tabular}
\\

\begin{tabular}[c]{@{}l@{}}Tumor grade\\prediction\end{tabular}
& 0.29
& \begin{tabular}[c]{@{}c@{}}0.77\\{[0.71,0.83]}\end{tabular}
& \begin{tabular}[c]{@{}c@{}}0.51\\{[0.41,0.65]}\end{tabular}
& \begin{tabular}[c]{@{}c@{}}1.77\\{[1.48,2.23]}\end{tabular}
& \begin{tabular}[c]{@{}c@{}}0.13\\{[0.09,0.17]}\end{tabular}
\\ \hline

\end{tabular}
\endgroup
\end{table}

\section{Discussion}

\gls{uq} is increasingly recognized as a key requirement for trustworthy artificial intelligence in medical imaging, yet its practical value remains highly dependent on the task and application context. In this work, we performed a task-aware evaluation of uncertainty within a multi-task \gls{dl} framework for glioma diagnosis, analyzing its \gls{mc} sample convergence, magnitude and decomposition, calibration, and operational utility, as well as its interactions between tumor segmentation and tumor molecular subtyping features. Collectively, our results provide insights into both the capabilities and limitations of uncertainty estimates in this clinical application.

We first assessed the convergence of average uncertainty estimates as a function of the number of \gls{mc} samples and dropout rate. Across tasks, mean uncertainty values generally reached a plateau after approximately 20--30 \gls{mc} samples, indicating that further sampling produced limited changes in average uncertainty magnitude. This supports the use of a computationally practical number of \gls{mc} samples for the subsequent analyses.

Increasing the dropout rate generally led to higher uncertainty magnitudes, reflecting broader posterior approximations and increased predictive variability. However, excessive dropout resulted in reduced predictive performance and degraded uncertainty quality, suggesting that overly strong stochastic perturbations can degrade predictive outputs and reduce the practical value of uncertainty estimates. This highlights the importance of balancing stochasticity and predictive fidelity when computing uncertainty through \gls{mcd}.

The analysis of uncertainty decomposition revealed consistent task-dependent patterns. Aleatoric uncertainty dominated in classification tasks, whereas epistemic uncertainty was more prominent in tumor segmentation, which likely reflects fundamental differences in the nature of the tasks. Molecular subtype prediction from \gls{mri} is inherently limited by the indirect relationship between imaging appearance and genetic status, introducing irreducible uncertainty possibly due to biological variability. In contrast, tumor segmentation is more directly informed by image appearance, tumor boundaries, and local texture patterns. Therefore, higher epistemic uncertainty in segmentation may reflect model-related limitations in representing atypical morphologies, ambiguous boundaries, or underrepresented imaging patterns in \gls{mri}.

Importantly, however, uncertainty magnitude alone did not directly indicate operational utility. Components that dominated in magnitude did not necessarily perform best in the evaluation metrics. Decomposing predictive uncertainty into aleatoric and epistemic components did not consistently improve error detection performance either. This finding is consistent with the mathematical definition of the decomposition used in this work: predictive uncertainty can be high because of aleatoric uncertainty, epistemic uncertainty, or both. Therefore, when one uncertainty source dominates, predictive uncertainty may already provide a strong operational signal for error detection, even if it does not distinguish why the model is uncertain. The bootstrap confidence intervals supported this interpretation, showing that the operational utility of predictive uncertainty was comparable to that of its components across tasks.


These findings are in line with a recent work by \cite{kahl2024values}, which questions the practical utility of uncertainty decomposition in real-world settings by demonstrating that uncertainty decomposition in segmentation works in synthetic settings but does not necessarily translate to real-world data, and therefore the benefit or feasibility of uncertainty decomposition cannot be assumed a priori. Instead, its value must be evaluated empirically for each specific task and dataset. Our results extend these observations to a multi-task setting including three classification tasks and demonstrate that uncertainty decomposition provides insight into the origin of predictive uncertainty but its operational utility is task-limited.

Calibration and quality analysis further demonstrated that the quality of uncertainty estimates was dependent on the amount of dropout applied. Moderate dropout levels achieved the best performance, while excessive dropout degraded calibration and predictive likelihood. This likely reflects the trade-off between model confidence and predictive accuracy: insufficient stochasticity leads to overconfident predictions, whereas excessive stochasticity reduces predictive consistency. These results highlight that uncertainty quality cannot be optimized independently of predictive performance, reinforcing the need for joint evaluation of both aspects when developing diagnostic AI systems.

The behavior of ECE, NLL, and AUC across dropout rates further illustrates that calibration and predictive performance capture related but distinct properties of the model probability outputs. ECE evaluates calibration by comparing predicted confidence with empirical accuracy, and therefore reflects whether the numerical confidence values are aligned with the observed correctness rate. In contrast, NLL evaluates the probability assigned to the true class, averaged over all cases, and therefore depends on both confidence and correctness. It penalizes probability mass assigned away from the true class, especially for confident incorrect predictions. Conversely, AUC evaluates discriminative performance by measuring how well the model separates cases across different classification thresholds, without directly assessing whether the predicted probabilities are calibrated.

These differences explain why the metrics did not always change in parallel across dropout rates. A decrease in ECE at a specific dropout configuration can indicate improved agreement between confidence and empirical accuracy, without necessarily implying that the model assigns higher probabilities to the true classes or improves class discrimination. Conversely, an increase in NLL can occur when the average probability assigned to the true class decreases, even if the confidence values remain reasonably aligned with empirical accuracy. Similarly, AUC may decrease when the separation of classes worsens, even if the calibration error is unchanged or reduced. These observations support evaluating ECE, NLL, and AUC, together, rather than interpreting any single metric as a complete summary of the model predictive performance.

In the tumor segmentation task, case-level uncertainty was negatively correlated with segmentation accuracy, indicating that uncertainty estimates reflected meaningful variations in prediction quality. The strength of this relationship depended strongly on the spatial aggregation strategy. Full-brain aggregation showed weak associations with \gls{dsc}, suggesting that averaging uncertainty over the entire brain is not well suited for case-level segmentation reliability assessment. This is likely because most brain voxels correspond to confidently predicted background and contribute limited information about tumor delineation quality, thereby diluting the uncertainty signal arising from the lesion region. In contrast, aggregation within the predicted tumor region showed the strongest and most stable negative association with segmentation accuracy.

The dilated tumor and boundary-weighted analyses further addressed whether uncertainty immediately outside the predicted boundary could provide additional information. These strategies also showed negative associations with \gls{dsc}, supporting the relevance of spatially localized uncertainty around the tumor, although they did not provide a clearer advantage over predicted tumor aggregation and introduced additional hyperparameters such as dilation radius or distance-weighting scale.

A central objective of this study was to evaluate the operational utility of uncertainty estimates by quantifying their ability to detect erroneous predictions. Across tasks, uncertainty estimates identified errors better than expected from the baseline error rates, with Lift values above 1 and \gls{aurc} values indicating reduced residual risk among retained high-confidence cases. These results indicate that uncertainty estimates can serve as practical indicators of prediction reliability in the studied AI-based glioma diagnosis framework.

The comparison with \gls{de} and \gls{mcde} further showed that this operational utility was not specific to \gls{mcd}. Across tasks, ensemble-based uncertainty estimates achieved performance comparable to \gls{mcd}, although no method consistently dominated across all metrics. This supports the generality of the task-aware evaluation framework, while also showing that the preferred \gls{uq} method may depend on the intended use case: ranking errors, identifying rare failures, or reducing residual risk among retained cases.

For tumor segmentation, the binary definition of segmentation error depended on the selected \gls{dsc} threshold. The threshold-sensitivity analysis showed that the segmentation error rate increased, as expected, when stricter \gls{dsc} thresholds were used. Nevertheless, segmentation uncertainty remained informative across the evaluated thresholds, with high error-detection performance and Lift values above 1. This indicates that the operational conclusion that segmentation uncertainty identifies low-quality segmentations was not driven by the primary threshold of $\mathrm{DSC}\leq0.70$, although the absolute values of AP and Lift varied with the induced error prevalence.

Because the classification tasks rely on shared representations learned jointly with segmentation, we further investigated their interactions. Tumor segmentation quality was significantly associated with classification correctness for \gls{idh} mutation status and tumor grade prediction, indicating that errors in structural representation learning may propagate to classification tasks. In contrast, the association between segmentation quality and 1p/19q co-deletion status prediction did not reach statistical significance. This suggests that, while accurate structural representation is important for certain molecular predictions, its influence may vary depending on the specific biological feature being predicted. One possible explanation is that imaging correlates of 1p/19q co-deletion status are less strongly linked to tumor morphology captured by segmentation, and may instead depend more on subtler texture or intensity patterns. Overall, these findings support the hypothesis that tumor segmentation performance partially reflects the quality of shared latent features used for molecular prediction, but that this dependency is task-specific rather than universal.

Correlation analysis revealed moderate associations between tumor segmentation and classification uncertainty for predictive and aleatoric components, suggesting that some uncertainty patterns are shared across tasks. The bootstrap confidence intervals supported positive associations for predictive and aleatoric uncertainty across all three classification tasks. In contrast, epistemic uncertainty showed weaker and less consistent relationships: its confidence intervals included zero for \gls{idh} mutation status and tumor grade prediction, while for 1p/19q co-deletion status prediction the association was weakly negative. Thus, shared cross-task uncertainty patterns were mainly observed for predictive and aleatoric uncertainty, rather than for the epistemic component.

Tumor segmentation uncertainty alone demonstrated moderate ability to predict classification errors, with Lift values above 1 across all three classification tasks, indicating that it contained information about classification reliability. However, when comparing predictive uncertainty scores directly, classification-specific uncertainty achieved higher Lift and lower \gls{aurc} than tumor segmentation uncertainty for all three classification tasks. This suggests that although segmentation contributes to classification reliability, classification-specific predictive uncertainty remains the most informative predictor of classification error.

To investigate whether integrating information across tasks could improve reliability assessment, we proposed a composite trust score combining classification and tumor segmentation uncertainty. The trust score identified low-trust cases with higher classification error rates than expected under random selection across tasks, with Lift values above 1. However, compared with predictive classification uncertainty alone, it did not provide a consistent improvement. For 1p/19q co-deletion status prediction, the trust score improved the point estimates for \gls{u-auc}, AP, Lift, and \gls{aurc}. This improvement was not observed for \gls{idh} mutation status or tumor grade prediction, where predictive classification uncertainty performed comparably or better depending on the metric. This result suggests that classification uncertainty already captures the most relevant information about classification reliability in this setting.

The previous finding highlights an important principle for trustworthy AI development: additional complexity does not necessarily improve reliability estimation. Instead, task-specific predictive uncertainty measures may provide the most direct and informative indicators of prediction confidence. At the same time, the trust score provides a useful conceptual framework for integrating uncertainty across tasks. Although its current formulation did not consistently improve selective risk reduction, it establishes a basis for future developments. Future trust-score formulations may require task-specific calibration, non-linear integration strategies, or explicit modeling of when tumor segmentation uncertainty is expected to contribute additional information beyond classification uncertainty alone.


This study has some limitations. First, uncertainty was evaluated using structural \gls{mri} modalities only. While these sequences form the backbone of current radiological assessment, advanced imaging techniques such as diffusion- and perfusion-weighted \gls{mri} provide complementary information that may be of utility to strengthen interpretability and robustness of the uncertainty estimates.

Second, our convergence analysis focused on average uncertainty magnitudes as a function of the number of \gls{mc} samples and dropout rate. Therefore, it should not be interpreted as a complete stability assessment. We did not evaluate case-wise ranking stability, repeatability across independently trained models or random seeds, or the variability of downstream error-detection metrics across repeated \gls{mc} realizations. These aspects represent important directions for future work, particularly when uncertainty estimates are used for patient-level referral or prioritization.

Beyond the specific methods evaluated in this study, the task-aware evaluation framework proposed here provides a general basis for assessing uncertainty estimates in clinically oriented \gls{dl} models. Because the framework evaluates convergence, calibration, uncertainty decomposition, error-detection performance, and selective prediction, its principles can be applied to newer \gls{uq} methods beyond \gls{mcd}, \gls{de}, and \gls{mcde}. Future work may therefore extend this comparison to additional \gls{uq} approaches and assess method-specific trade-offs in calibration, computational cost, scalability, and clinical deployment feasibility.

\section{Conclusion}

This study presents a comprehensive, task-aware evaluation of \gls{uq} in a multi-task \gls{dl} framework for glioma diagnosis. 
Our findings show that average \gls{mcd} uncertainty estimates converged with a practical number of \gls{mc} samples, that calibration and uncertainty quality depended on the dropout configuration, and that uncertainty estimates provided meaningful information about model reliability. Across tasks, uncertainty estimates enabled identification of prediction errors and supported selective-prediction analyses, although their utility depended on the task, uncertainty component, and operational metric considered.

Uncertainty decomposition offered insight into task-specific uncertainty sources, with aleatoric uncertainty being more prominent in classification and epistemic uncertainty being more prominent in tumor segmentation. However, decomposition did not provide a consistent operational advantage over predictive uncertainty alone. For tumor segmentation, uncertainty estimates were informative for identifying low-quality segmentations, particularly when aggregated within tumor-focused regions rather than across the full brain.

Although tumor segmentation performance and uncertainty were associated with classification reliability, classification predictive uncertainty remained the most direct and informative indicator of classification error, and the proposed composite trust score did not consistently improve over it. The comparison of \gls{mcd} with \gls{de} and \gls{mcde} further showed that the operational value of uncertainty estimates was not restricted to a single \gls{uq} method. Overall, these findings emphasize that \gls{uq} should be evaluated in relation to the intended clinical task and use case. The task-aware evaluation framework proposed here provides practical guidance for assessing uncertainty in \gls{dl}-based diagnostic tools and represents a promising step towards trustworthy AI for glioma diagnosis.

\acks{This work is part of the “Trustworthy AI for \gls{mri}” ICAI lab within the project ROBUST: Trustworthy AI-based Systems for Sustainable Growth with project number \linebreak KICH3.LTP.20.006, financed by the Dutch Research Council (NWO), GE HealthCare, and the Dutch Ministry of Economic Affairs and Climate Policy (EZK) under the program LTP KIC 2020-2023.}

%
\ethics{The work follows appropriate ethical standards in conducting research and writing the manuscript, following all applicable laws and regulations regarding treatment of animals or human subjects.}

\coi{Authors GEMR, MS, and SKlein are part of the Erasmus MC ICAI lab "Trustworthy AI for MRI", a public-private research program partially funded by GE HealthCare (payment to institution). CP and SKaushik are employees of GE HealthCare. SvdV declares no competing interests.}

\data{
The public datasets included in this study are available online:

\begin{itemize}[leftmargin=*, itemsep=2pt, topsep=4pt, parsep=0pt]
  \item BraTS: \url{http://braintumorsegmentation.org/}
  \item Brain Tumor Progression: \url{https://doi.org/10.7937/K9/TCIA.2018.15quzvnb}
  \item CPTAC-GBM: \url{https://doi.org/10.7937/K9/TCIA.2018.3rje41q1}
  \item EGD: \url{https://xnat.bmia.nl/REST/projects/egd}
  \item IvyGAP: \url{https://doi.org/10.7937/K9/TCIA.2016.XLwaN6nL}
  \item REMBRANDT: \url{https://doi.org/10.7937/K9/TCIA.2015.588OZUZB}
  \item TCGA-GBM: \url{https://doi.org/10.7937/K9/TCIA.2016.RNYFUYE9}
  \item TCGA-LGG: \url{https://doi.org/10.7937/K9/TCIA.2016.L4LTD3TK}
\end{itemize}

The code used for this paper is publicly available in the following repository: \url{https://github.com/ErasmusMC-NeuroOnco/PrognosAIs-UQ}

}

\bibliography{references}


\clearpage
\appendix
\renewcommand*{\theHsection}{appendix.\Alph{section}}
\renewcommand*{\theHsubsection}{\theHsection.\arabic{subsection}}
\section{Implementation details}
\subsection{Pre-processing}
\label{app:pre-processing}

Prior to model training, all \gls{mri} scans underwent a standardized pre-processing pipeline consistent with common practices in Medical Image Analysis. These steps were designed to reduce inter-subject variability, improve anatomical alignment across patients, and ensure numerical stability during network optimization.

First, all scans were spatially registered to the MNI152 atlas space \citep{fonov2009unbiased, fonov2011unbiased} to establish a common anatomical reference frame. This atlas has a resolution of 1×1×1 mm$^3$ and a size of 197×233×189
voxels. Modality-specific registration was performed to preserve anatomical correspondence: \gls{t1w} and \gls{t1wce} images were registered to the \gls{t1w} MNI152 atlas, while \gls{t2w} and \gls{flair} images were registered to the \gls{t2w} atlas template. This modality-aware registration strategy ensures optimal alignment of tissue contrasts while minimizing interpolation artifacts across sequences. Standard-space alignment reduces anatomical variability unrelated to pathology and facilitates consistent learning of spatial patterns by the \gls{nn}.

Following registration, bias field correction was applied to each scan to mitigate low-frequency intensity inhomogeneities caused by magnetic field non-uniformities. Correcting these intensity distortions is particularly important in multi-center datasets, like the one used in this work, as scanner-dependent intensity variations can otherwise introduce spurious features that degrade model generalization.

Subsequently, skull-stripping was performed using the HD-BET tool \citep{isensee2019automated}  to generate an accurate brain mask for the MNI152 atlas. Removing non-brain tissues (e.g., skull, fat, and extracranial structures) serves two purposes: (i) it restricts the learning process to anatomically relevant regions and (ii) reduces the risk of the model exploiting non-biological artifacts. The resulting brain mask was further used to compute a tight bounding box around the intracranial volume, allowing spatial cropping of the scans, and resulting in a uniform scan size of 145x182x152 voxels. This step reduces computational burden, decreases background redundancy, and increases the effective proportion of informative voxels presented to the model.

Finally, intensity normalization was performed via z-score standardization within the brain mask region. For each scan, voxel intensities were centered and scaled using the mean and standard deviation computed over brain tissue only. Masked normalization avoids contamination from background voxels and ensures comparable intensity distributions across subjects and modalities, thereby promoting stable training dynamics and improving convergence.

\hypersetup{bookmarksdepth=1}
\subsection{\texorpdfstring{\protect\hypertarget{appendix.architecture.heading}{Architecture and training procedure}}{Architecture and training procedure}}
\begin{figure*}[h]
    \centering
    \includegraphics[width=1\linewidth, height=0.413\textheight]{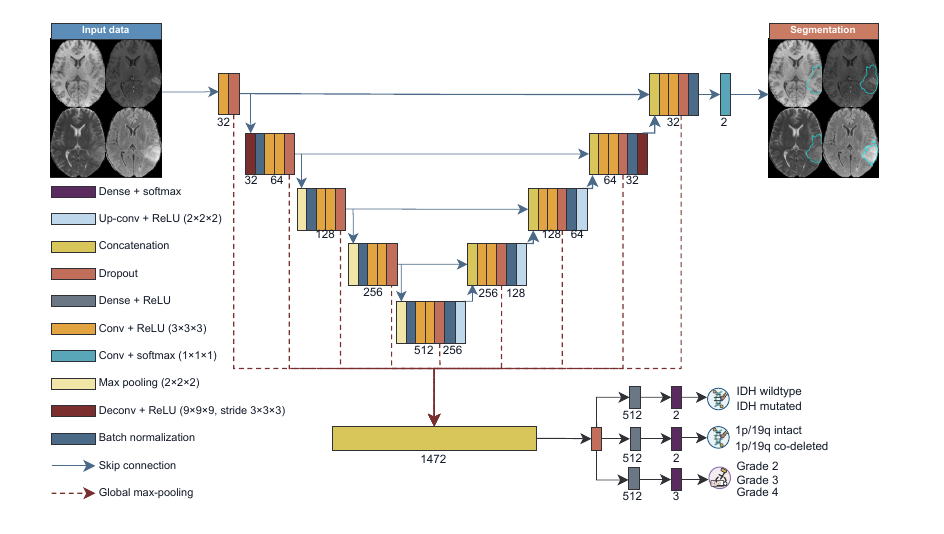}
      \caption{Overview of the multi-task \gls{dl} network used for the tumor segmentation and prediction of IDH mutation status, 1p/19q co-deletion status and tumor grade. The architecture processes full 3D volumes. The labels in the figure define the layer type, and the numbers indicate either the number of filters, dense units or features. Figure based on the one presented in the work of \cite{van2023combined}.}
  \label{fig:architecture}
\end{figure*}

\hypersetup{bookmarksdepth=2}
\bookmark[dest=appendix.architecture.heading,level=2]{Architecture and training procedure}
\makeatletter\def\@currentHref{appendix.architecture.heading}\makeatother
\label{app:architecture}
The architecture \citep{van2023combined,ronneberger2015u} depicted in Figure \ref{fig:architecture} processes full 3D volumes. It was trained for up to 100 epochs with ADAM \citep{kingma2014adam} optimizer and minimizing the following weighted loss function \citep{van2023combined}: 

\begin{equation}
\begin{aligned}
\mathcal{L} ={}&
W_{\text{\gls{idh}}} \cdot \mathcal{L}_{\text{\gls{idh}}}
+ W_{\text{1p/19q}} \cdot \mathcal{L}_{\text{1p/19q}} \\
&+ W_{\text{grade}} \cdot \mathcal{L}_{\text{grade}}
+ W_{\text{seg}} \cdot \mathcal{L}_{\text{seg}} .
\end{aligned}
\end{equation}
\begin{equation}
W_t = 
\begin{cases}
\frac{N_{\text{train}}}{N_t} & \text{for } t \in \{\text{\gls{idh}}, \text{1p/19q}, \text{grade}\} \\
1 & \text{for } t = \text{seg}
\end{cases},
\end{equation}

\noindent
where $N_{train}$ corresponds to the number of training subjects, and $N_t$ to the amount of available labeled samples for each task $t$.

To account for missing data and intrinsic class imbalance in each task, a masked, class-weighted categorical cross-entropy loss was used. This formulation ensures that only samples with available labels contribute to the loss, and that underrepresented classes are not neglected during optimization. Let a batch contain \( N \) samples and \( C \) classes. For each sample \( i \in \{1, \dots, N\} \), let \( \mathbf{y}_i = (y_{i,1}, \dots, y_{i,C}) \) denote the one-hot encoded ground truth label, and \( \hat{\mathbf{y}}_i = (\hat{y}_{i,1}, \dots, \hat{y}_{i,C}) \) the predicted class probabilities. Let \( m_i \in \{0,1\} \) be a binary mask indicating whether sample \( i \) has a valid label. A fixed weight \( w_c \) is assigned to each class \( c \), computed from the full training dataset as:

\begin{equation}
    w_c = \frac{N_t}{C \cdot n_c},
\end{equation}

\noindent
where \( N_t \) corresponds to the amount of labeled training samples for the task, and \( n_c \) is the number of occurrences of class \( c \) among those training samples. The masked, class-weighted loss for a batch is then defined as \citep{van2023combined}:

\begin{equation}
\mathcal{L}_{\text{masked}} =
\begin{cases}
\dfrac{1}{\sum_{i=1}^N m_i}
\displaystyle
\sum_{i=1}^N m_i \,
\Bigl(
- \sum_{c=1}^C
w_c \, y_{i,c}
\log \hat{y}_{i,c}
\Bigr),
\\[6pt]
\hfill \text{if } \sum_{i=1}^N m_i > 0,
\\[8pt]
0, \quad \text{otherwise}.
\end{cases}
\end{equation}

For the tumor segmentation task, we use the \gls{dsc} loss, which directly optimizes for the overlap between the predicted segmentation and the ground truth. Let \( \hat{p}_j \in [0,1] \) and \( g_j \in [0,1] \) denote the predicted and ground truth values for voxel \( j \) over a total of \( M \) voxels. The \gls{dsc} loss is defined as:

\[
\mathcal{L}_{\text{\gls{dsc}}} = 1 - \frac{2 \sum_{j=1}^M \hat{p}_j g_j }{\sum_{j=1}^M \hat{p}_j^2 + \sum_{j=1}^M g_j^2 + \epsilon},
\]
\noindent
where \( \epsilon \) is a small constant added for numerical stability.

\section{Data}
\label{ap2:data}

Table~\ref{tab:data-detailed-overview} provides a detailed overview of the dataset used in this study. The table reports the number of cases in the training and test sets, together with the distribution of available labels for \gls{idh} mutation status, 1p/19q co-deletion status, and tumor grade. For each task, cases with missing annotations are reported as N/A, reflecting incomplete molecular or histopathological information. This breakdown highlights both the class imbalance across tasks and the presence of missing labels, which motivated the use of task-specific masking during training.

\section{Segmentation error threshold sensitivity analysis}
\label{app:segmentation-sensitivity-analysis}

To assess whether the operational utility of segmentation uncertainty depended on the binary definition of segmentation error, we repeated the segmentation error-detection analysis for three \gls{dsc} thresholds: $\tau_{\mathrm{DSC}}=0.60$, $\tau_{\mathrm{DSC}}=0.70$, and $\tau_{\mathrm{DSC}}=0.80$. These thresholds were chosen to represent a more lenient definition focused on severe failures, the primary intermediate operating point used in the main analysis, and a stricter definition of low-quality segmentation. For each threshold, segmentation error was defined as $\mathrm{DSC}\leq\tau_{\mathrm{DSC}}$, and the threshold-specific error rate, \gls{u-auc}, AP, and Lift were recomputed for predictive, aleatoric and epistemic uncertainty. This analysis was not used to optimize the segmentation error threshold, but to assess whether the conclusion that segmentation uncertainty identifies low-quality segmentations was robust to the selected binarization.

\begin{table}[t]
\centering
\begingroup

\caption{Sensitivity of segmentation uncertainty error detection to the \gls{dsc} threshold used to define segmentation error. For each threshold, segmentation error was defined as $\mathrm{DSC}\leq\tau_{\mathrm{DSC}}$. Error rate is reported as the proportion of cases classified as segmentation errors. \gls{u-auc}, AP, and Lift are reported for predictive, aleatoric, and epistemic uncertainty, with $95\%$ confidence intervals obtained by $1000\times$ bootstrap resampling of the test set. Arrows in metric column headers indicate whether higher ($\uparrow$) or lower ($\downarrow$) values are preferable. }
\label{tab:segmentation_threshold_sensitivity}
\scriptsize
\setlength{\tabcolsep}{2pt}
\begin{tabular}{cclccc}
\hline
$\tau_{\mathrm{DSC}}$
& \begin{tabular}[c]{@{}c@{}}Error rate $\downarrow$\end{tabular}
& \multicolumn{1}{c}{Uncertainty}
& \begin{tabular}[c]{@{}c@{}}U-AUC $\uparrow$\end{tabular}
& AP $\uparrow$
& Lift $\uparrow$
\\ \hline

&
& Predictive
& \begin{tabular}[c]{@{}c@{}}0.96\\{[0.93,0.98]}\end{tabular}
& \begin{tabular}[c]{@{}c@{}}0.78\\{[0.64,0.90]}\end{tabular}
& \begin{tabular}[c]{@{}c@{}}7.96\\{[5.73,12.41]}\end{tabular}
\\

0.60
& 0.10
& Aleatoric
& \begin{tabular}[c]{@{}c@{}}0.97\\{[0.95,0.99]}\end{tabular}
& \begin{tabular}[c]{@{}c@{}}0.82\\{[0.67,0.92]}\end{tabular}
& \begin{tabular}[c]{@{}c@{}}8.33\\{[5.99,13.09]}\end{tabular}
\\

&
& Epistemic
& \begin{tabular}[c]{@{}c@{}}0.96\\{[0.93,0.98]}\end{tabular}
& \begin{tabular}[c]{@{}c@{}}0.76\\{[0.61,0.88]}\end{tabular}
& \begin{tabular}[c]{@{}c@{}}7.77\\{[5.61,12.26]}\end{tabular}
\\

\hline

&
& Predictive
& \begin{tabular}[c]{@{}c@{}}0.95\\{[0.92,0.98]}\end{tabular}
& \begin{tabular}[c]{@{}c@{}}0.82\\{[0.72,0.91]}\end{tabular}
& \begin{tabular}[c]{@{}c@{}}4.95\\{[3.89,6.67]}\end{tabular}
\\

0.70
& 0.17
& Aleatoric
& \begin{tabular}[c]{@{}c@{}}0.96\\{[0.94,0.98]}\end{tabular}
& \begin{tabular}[c]{@{}c@{}}0.83\\{[0.72,0.91]}\end{tabular}
& \begin{tabular}[c]{@{}c@{}}5.01\\{[3.93,6.80]}\end{tabular}
\\

&
& Epistemic
& \begin{tabular}[c]{@{}c@{}}0.95\\{[0.92,0.98]}\end{tabular}
& \begin{tabular}[c]{@{}c@{}}0.81\\{[0.71,0.90]}\end{tabular}
& \begin{tabular}[c]{@{}c@{}}4.90\\{[3.85,6.64]}\end{tabular}
\\

\hline

&
& Predictive
& \begin{tabular}[c]{@{}c@{}}0.93\\{[0.89,0.96]}\end{tabular}
& \begin{tabular}[c]{@{}c@{}}0.88\\{[0.81,0.93]}\end{tabular}
& \begin{tabular}[c]{@{}c@{}}2.83\\{[2.41,3.46]}\end{tabular}
\\

0.80
& 0.31
& Aleatoric
& \begin{tabular}[c]{@{}c@{}}0.93\\{[0.89,0.97]}\end{tabular}
& \begin{tabular}[c]{@{}c@{}}0.89\\{[0.83,0.95]}\end{tabular}
& \begin{tabular}[c]{@{}c@{}}2.87\\{[2.45,3.53]}\end{tabular}
\\

&
& Epistemic
& \begin{tabular}[c]{@{}c@{}}0.92\\{[0.88,0.95]}\end{tabular}
& \begin{tabular}[c]{@{}c@{}}0.87\\{[0.80,0.92]}\end{tabular}
& \begin{tabular}[c]{@{}c@{}}2.80\\{[2.38,3.42]}\end{tabular}
\\

\hline
\end{tabular}
\endgroup
\end{table}

As expected, increasing the \gls{dsc} threshold resulted in a higher segmentation error rate, from 0.1 at $\tau_{\mathrm{DSC}}=0.60$, to 0.17 at $\tau_{\mathrm{DSC}}=0.70$, and 0.31 at $\tau_{\mathrm{DSC}}=0.80$. Despite this change in error prevalence, segmentation uncertainty remained strongly informative across thresholds. For predictive uncertainty, \gls{u-auc} remained high across all thresholds, ranging from 0.96 at $\tau_{\mathrm{DSC}}=0.60$ to 0.93 at $\tau_{\mathrm{DSC}}=0.80$. AP values were consistently above the corresponding threshold-specific baseline error rates, and Lift values remained greater than 1 for all thresholds. Similar patterns were observed for the aleatoric and epistemic uncertainty components. These results indicate that the operational conclusion that segmentation uncertainty identifies low-quality segmentations was not driven by the specific primary threshold of $\mathrm{DSC}\leq0.70$, although the absolute metric values varied with the prevalence of segmentation errors induced by each threshold.

\hypersetup{bookmarksdepth=0}
\section{\texorpdfstring{\protect\hypertarget{appendix.correlation.heading}{Correlation patterns between segmentation uncertainty estimates and tumor segmentation DSC}}{Correlation patterns between segmentation uncertainty estimates and tumor segmentation DSC}}
\hypersetup{bookmarksdepth=2}
\bookmark[dest=appendix.correlation.heading,level=1]{Correlation patterns between segmentation uncertainty estimates and tumor segmentation DSC}
\makeatletter\def\@currentHref{appendix.correlation.heading}\makeatother
\label{app:segmentation-correlation}

\begin{figure}[H]
    \centering
    \includegraphics[width=0.93\linewidth]{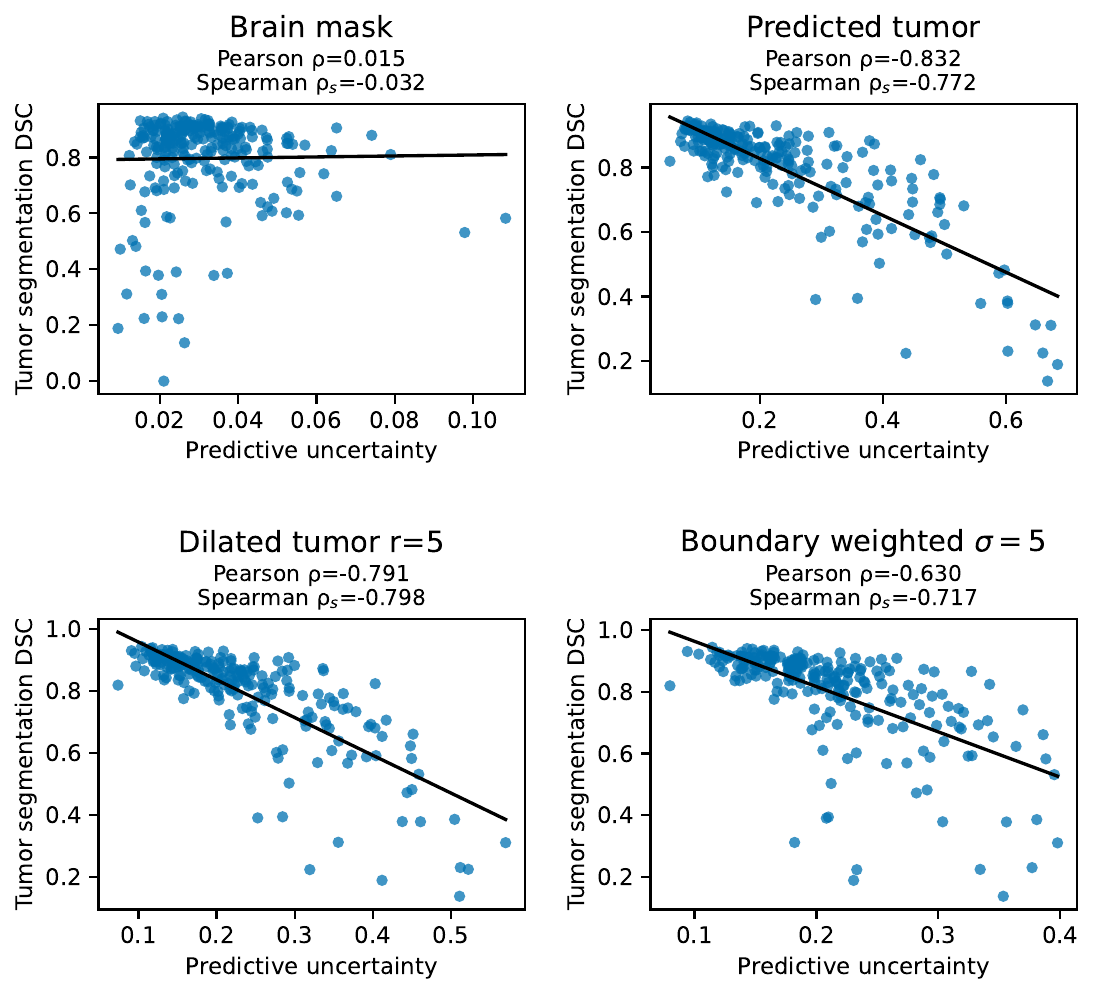}
    
    \caption{Case-wise association between predictive segmentation uncertainty and tumor segmentation DSC at the selected MCD configuration ($T=30$, $\delta=0.25$). Results are shown for brain mask, predicted tumor, 5-voxel dilated tumor, and boundary-weighted aggregation ($\sigma=5$ voxels). Each point represents one test case; Pearson $\rho$ and Spearman $\rho_s$ are reported for each strategy.}
    \label{fig:segmentation-uncertainty-scatter}
    
\end{figure}

Figure~\ref{fig:segmentation-uncertainty-scatter} shows the case-wise relationship between predictive segmentation uncertainty and tumor segmentation \gls{dsc} at the selected \gls{mcd} configuration. This analysis was included to assess whether the uncertainty--\gls{dsc} associations reported using Pearson correlation were consistent with the corresponding rank-based Spearman correlation.

Across the tumor-localized uncertainty scores, Pearson and Spearman correlations showed the same overall interpretation: higher predictive uncertainty was associated with lower tumor segmentation \gls{dsc}. For predicted tumor aggregation, both coefficients indicated a strong negative association, with Pearson $\rho=-0.832$ and Spearman $\rho_s=-0.772$. A similar pattern was observed for dilated tumor aggregation, with Pearson $\rho=-0.791$ and Spearman $\rho_s=-0.798$. Boundary-weighted aggregation also showed a negative relationship, with Pearson $\rho=-0.630$ and Spearman $\rho_s=-0.717$.

For brain mask aggregation, both coefficients were close to zero, with Pearson $\rho=0.015$ and Spearman $\rho_s=-0.032$, indicating no meaningful association between brain-level predictive uncertainty and tumor segmentation \gls{dsc}. Thus, the conclusions were consistent across Pearson and Spearman correlation: tumor-localized uncertainty showed a negative uncertainty--performance relationship, whereas brain-level aggregation did not. This supports the use of Pearson correlation in the main analysis, while showing that the observed findings were not dependent on assuming a strictly linear relationship.

\begin{table*}[h]
\centering
\caption{Detailed distribution of the data used for training and testing the Deep Learning (DL) architecture for each one of the classification tasks. N/A (not available) represents missing data. Tumor segmentation ground truth was available for all cases.}
\label{tab:data-detailed-overview}
\begin{tabular}{cccccccccccc}
\hline
\multirow{2}{*}{Dataset} & \multirow{2}{*}{Subset} & \multicolumn{3}{c}{IDH mutation status} & \multicolumn{3}{c}{1p/19q co-deletion status} & \multicolumn{4}{c}{Tumor grade} \\ \cline{3-12} 
 &  & Wildtype & Mutated & N/A & Intact & Co-deleted & N/A & 2 & 3 & 4 & N/A \\ \hline
\multirow{7}{*}{Train} & BraTS & 0 & 0 & 156 & 0 & 0 & 156 & 0 & 0 & 0 & 156 \\
 & \begin{tabular}[c]{@{}c@{}}Brain tumor\\ Progression\end{tabular} & 0 & 0 & 20 & 0 & 0 & 20 & 0 & 0 & 0 & 20 \\
 & CPTAC-GBM & 0 & 0 & 45 & 0 & 0 & 45 & 0 & 0 & 0 & 45 \\
 & EGD & 312 & 155 & 308 & 186 & 73 & 516 & 135 & 80 & 502 & 58 \\
 & IvyGAP & 32 & 6 & 1 & 27 & 3 & 9 & 0 & 1 & 36 & 2 \\
 & In-house & 86 & 53 & 183 & 107 & 22 & 193 & 24 & 6 & 191 & 101 \\
 & REMBRANDT & 0 & 0 & 109 & 0 & 0 & 109 & 37 & 21 & 32 & 19 \\ \hline
\multirow{2}{*}{Test} & TCGA-GBM & 107 & 5 & 21 & 127 & 0 & 6 & 0 & 0 & 132 & 1 \\
 & TCGA-LGG & 22 & 80 & 1 & 78 & 25 & 0 & 45 & 58 & 0 & 0 \\ \hline
\end{tabular}
\end{table*}

\section{Comparison of \texorpdfstring{\gls{uq}}{UQ} Methods}
\label{app:comparison-uq}

Table~\ref{tab:uq_method_components_comparison_appendix} provides the component-wise comparison of \gls{mcd}, \gls{de}, and \gls{mcde} using aleatoric and epistemic uncertainty. These analyses complement the main predictive-uncertainty comparison by showing how the decomposed uncertainty components behave across tasks and methods. For tumor segmentation, uncertainty was aggregated within the predicted tumor region, matching the strategy used in the main method-comparison analysis.

Overall, aleatoric uncertainty showed trends broadly consistent with predictive uncertainty, with useful error-detection performance across classification tasks and high performance for tumor segmentation. 
Epistemic uncertainty was more variable across methods and tasks. It remained informative for tumor segmentation with all three methods, with the strongest \gls{u-auc}, \gls{ap}, and Lift observed for MCD, but was generally weaker for classification error detection.

\begin{table*}[t]
\centering
\begingroup

\scriptsize
\setlength{\tabcolsep}{1.8pt}
\caption{Comparison of Uncertainty Quantification methods using aleatoric and epistemic uncertainty. Assessment is made using Uncertainty-based Receiver Operating Characteristic Area Under the Curve (U-AUC), Average Precision (AP), Lift, and Area Under the Risk--Coverage Curve (\gls{aurc}), derived from using each uncertainty component as a predictor of errors. Values are reported with $95\%$ confidence intervals obtained by $1000\times$ bootstrap resampling of the test set. The error column denotes the method-specific classification error rate for classification tasks and the proportion of cases with DSC $\leq 0.70$ for tumor segmentation. For the latter, uncertainty was aggregated within the predicted tumor region. The highest displayed values for U-AUC, AP, and Lift, and the lowest displayed values for \gls{aurc}, are indicated in bold for each task and uncertainty component. Arrows in metric column headers indicate whether higher ($\uparrow$) or lower ($\downarrow$) values are preferable.}
\label{tab:uq_method_components_comparison_appendix}

\resizebox{\textwidth}{!}{
\begin{tabular}{llc|cccc|cccc}
\hline
\multirow{2}{*}{Task} & \multirow{2}{*}{Method} & \multirow{2}{*}{Error $\downarrow$}
& \multicolumn{4}{c|}{Aleatoric uncertainty}
& \multicolumn{4}{c}{Epistemic uncertainty} \\
\cline{4-11}
& & 
& U-AUC $\uparrow$ & AP $\uparrow$ & Lift $\uparrow$ & \gls{aurc} $\downarrow$
& U-AUC $\uparrow$ & AP $\uparrow$ & Lift $\uparrow$ & \gls{aurc} $\downarrow$ \\
\hline

\multirow{3}{*}{\begin{tabular}[c]{@{}l@{}}\gls{idh}\\mutation\end{tabular}}
& \gls{de}
& 0.17
& \begin{tabular}[c]{@{}c@{}}\textbf{0.73}\\{[0.62,0.82]}\end{tabular}
& \begin{tabular}[c]{@{}c@{}}\textbf{0.39}\\{[0.26,0.56]}\end{tabular}
& \begin{tabular}[c]{@{}c@{}}\textbf{2.26}\\{[1.66,3.31]}\end{tabular}
& \begin{tabular}[c]{@{}c@{}}\textbf{0.09}\\{[0.05,0.15]}\end{tabular}
& \begin{tabular}[c]{@{}c@{}}0.64\\{[0.54,0.74]}\end{tabular}
& \begin{tabular}[c]{@{}c@{}}0.28\\{[0.17,0.43]}\end{tabular}
& \begin{tabular}[c]{@{}c@{}}1.60\\{[1.20,2.45]}\end{tabular}
& \begin{tabular}[c]{@{}c@{}}0.12\\{[0.07,0.18]}\end{tabular}
\\

& \gls{mcd}
& 0.18
& \begin{tabular}[c]{@{}c@{}}0.71\\{[0.61,0.80]}\end{tabular}
& \begin{tabular}[c]{@{}c@{}}0.32\\{[0.22,0.46]}\end{tabular}
& \begin{tabular}[c]{@{}c@{}}1.79\\{[1.39,2.54]}\end{tabular}
& \begin{tabular}[c]{@{}c@{}}0.10\\{[0.05,0.15]}\end{tabular}
& \begin{tabular}[c]{@{}c@{}}\textbf{0.71}\\{[0.61,0.80]}\end{tabular}
& \begin{tabular}[c]{@{}c@{}}\textbf{0.34}\\{[0.23,0.50]}\end{tabular}
& \begin{tabular}[c]{@{}c@{}}\textbf{1.89}\\{[1.45,2.80]}\end{tabular}
& \begin{tabular}[c]{@{}c@{}}\textbf{0.10}\\{[0.05,0.16]}\end{tabular}
\\

& \gls{mcde}
& 0.17
& \begin{tabular}[c]{@{}c@{}}\textbf{0.73}\\{[0.62,0.82]}\end{tabular}
& \begin{tabular}[c]{@{}c@{}}0.36\\{[0.24,0.53]}\end{tabular}
& \begin{tabular}[c]{@{}c@{}}2.12\\{[1.55,3.13]}\end{tabular}
& \begin{tabular}[c]{@{}c@{}}\textbf{0.09}\\{[0.05,0.14]}\end{tabular}
& \begin{tabular}[c]{@{}c@{}}0.64\\{[0.54,0.73]}\end{tabular}
& \begin{tabular}[c]{@{}c@{}}0.30\\{[0.20,0.47]}\end{tabular}
& \begin{tabular}[c]{@{}c@{}}1.81\\{[1.31,2.83]}\end{tabular}
& \begin{tabular}[c]{@{}c@{}}0.12\\{[0.07,0.19]}\end{tabular}
\\
\hline

\multirow{3}{*}{\begin{tabular}[c]{@{}l@{}}1p/19q\\co-deletion\end{tabular}}
& \gls{de}
& 0.10
& \begin{tabular}[c]{@{}c@{}}0.83\\{[0.75,0.90]}\end{tabular}
& \begin{tabular}[c]{@{}c@{}}0.29\\{[0.18,0.49]}\end{tabular}
& \begin{tabular}[c]{@{}c@{}}2.95\\{[2.12,5.25]}\end{tabular}
& \begin{tabular}[c]{@{}c@{}}0.03\\{[0.01,0.04]}\end{tabular}
& \begin{tabular}[c]{@{}c@{}}0.66\\{[0.55,0.76]}\end{tabular}
& \begin{tabular}[c]{@{}c@{}}\textbf{0.18}\\{[0.10,0.34]}\end{tabular}
& \begin{tabular}[c]{@{}c@{}}\textbf{1.79}\\{[1.19,3.65]}\end{tabular}
& \begin{tabular}[c]{@{}c@{}}\textbf{0.05}\\{[0.03,0.08]}\end{tabular}
\\

& \gls{mcd}
& 0.10
& \begin{tabular}[c]{@{}c@{}}\textbf{0.84}\\{[0.77,0.91]}\end{tabular}
& \begin{tabular}[c]{@{}c@{}}\textbf{0.39}\\{[0.22,0.58]}\end{tabular}
& \begin{tabular}[c]{@{}c@{}}\textbf{3.70}\\{[2.51,6.19]}\end{tabular}
& \begin{tabular}[c]{@{}c@{}}0.03\\{[0.01,0.04]}\end{tabular}
& \begin{tabular}[c]{@{}c@{}}\textbf{0.68}\\{[0.58,0.77]}\end{tabular}
& \begin{tabular}[c]{@{}c@{}}0.16\\{[0.10,0.29]}\end{tabular}
& \begin{tabular}[c]{@{}c@{}}1.54\\{[1.22,2.59]}\end{tabular}
& \begin{tabular}[c]{@{}c@{}}\textbf{0.05}\\{[0.03,0.08]}\end{tabular}
\\

& \gls{mcde}
& 0.09
& \begin{tabular}[c]{@{}c@{}}0.81\\{[0.74,0.89]}\end{tabular}
& \begin{tabular}[c]{@{}c@{}}0.28\\{[0.17,0.47]}\end{tabular}
& \begin{tabular}[c]{@{}c@{}}3.02\\{[2.11,5.55]}\end{tabular}
& \begin{tabular}[c]{@{}c@{}}\textbf{0.02}\\{[0.01,0.04]}\end{tabular}
& \begin{tabular}[c]{@{}c@{}}0.60\\{[0.48,0.72]}\end{tabular}
& \begin{tabular}[c]{@{}c@{}}0.12\\{[0.08,0.22]}\end{tabular}
& \begin{tabular}[c]{@{}c@{}}1.33\\{[1.04,2.27]}\end{tabular}
& \begin{tabular}[c]{@{}c@{}}0.06\\{[0.03,0.09]}\end{tabular}
\\
\hline

\multirow{3}{*}{\begin{tabular}[c]{@{}l@{}}Tumor\\grade\end{tabular}}
& \gls{de}
& 0.30
& \begin{tabular}[c]{@{}c@{}}\textbf{0.80}\\{[0.73,0.85]}\end{tabular}
& \begin{tabular}[c]{@{}c@{}}\textbf{0.59}\\{[0.48,0.71]}\end{tabular}
& \begin{tabular}[c]{@{}c@{}}1.99\\{[1.65,2.47]}\end{tabular}
& \begin{tabular}[c]{@{}c@{}}0.13\\{[0.09,0.18]}\end{tabular}
& \begin{tabular}[c]{@{}c@{}}0.64\\{[0.56,0.72]}\end{tabular}
& \begin{tabular}[c]{@{}c@{}}\textbf{0.47}\\{[0.36,0.60]}\end{tabular}
& \begin{tabular}[c]{@{}c@{}}\textbf{1.58}\\{[1.31,2.02]}\end{tabular}
& \begin{tabular}[c]{@{}c@{}}0.22\\{[0.16,0.29]}\end{tabular}
\\

& \gls{mcd}
& 0.29
& \begin{tabular}[c]{@{}c@{}}0.79\\{[0.73,0.85]}\end{tabular}
& \begin{tabular}[c]{@{}c@{}}\textbf{0.59}\\{[0.47,0.72]}\end{tabular}
& \begin{tabular}[c]{@{}c@{}}\textbf{2.00}\\{[1.68,2.50]}\end{tabular}
& \begin{tabular}[c]{@{}c@{}}\textbf{0.12}\\{[0.09,0.17]}\end{tabular}
& \begin{tabular}[c]{@{}c@{}}\textbf{0.70}\\{[0.62,0.77]}\end{tabular}
& \begin{tabular}[c]{@{}c@{}}0.45\\{[0.35,0.58]}\end{tabular}
& \begin{tabular}[c]{@{}c@{}}1.52\\{[1.28,1.94]}\end{tabular}
& \begin{tabular}[c]{@{}c@{}}\textbf{0.17}\\{[0.12,0.24]}\end{tabular}
\\

& \gls{mcde}
& 0.28
& \begin{tabular}[c]{@{}c@{}}0.78\\{[0.71,0.84]}\end{tabular}
& \begin{tabular}[c]{@{}c@{}}0.53\\{[0.42,0.66]}\end{tabular}
& \begin{tabular}[c]{@{}c@{}}1.90\\{[1.58,2.40]}\end{tabular}
& \begin{tabular}[c]{@{}c@{}}\textbf{0.12}\\{[0.08,0.17]}\end{tabular}
& \begin{tabular}[c]{@{}c@{}}0.63\\{[0.55,0.71]}\end{tabular}
& \begin{tabular}[c]{@{}c@{}}0.41\\{[0.32,0.54]}\end{tabular}
& \begin{tabular}[c]{@{}c@{}}1.45\\{[1.21,1.90]}\end{tabular}
& \begin{tabular}[c]{@{}c@{}}0.21\\{[0.14,0.28]}\end{tabular}
\\
\hline

\multirow{3}{*}{\begin{tabular}[c]{@{}l@{}}Tumor\\segmentation\end{tabular}}
& \gls{de}
& 0.13
& \begin{tabular}[c]{@{}c@{}}\textbf{0.96}\\{[0.93,0.98]}\end{tabular}
& \begin{tabular}[c]{@{}c@{}}0.81\\{[0.66,0.90]}\end{tabular}
& \begin{tabular}[c]{@{}c@{}}6.11\\{[4.58,8.68]}\end{tabular}
& \begin{tabular}[c]{@{}c@{}}0.13\\{[0.12,0.13]}\end{tabular}
& \begin{tabular}[c]{@{}c@{}}0.93\\{[0.89,0.96]}\end{tabular}
& \begin{tabular}[c]{@{}c@{}}0.63\\{[0.46,0.80]}\end{tabular}
& \begin{tabular}[c]{@{}c@{}}4.75\\{[3.56,7.10]}\end{tabular}
& \begin{tabular}[c]{@{}c@{}}\textbf{0.12}\\{[0.12,0.13]}\end{tabular}
\\

& \gls{mcd}
& 0.17
& \begin{tabular}[c]{@{}c@{}}\textbf{0.96}\\{[0.94,0.98]}\end{tabular}
& \begin{tabular}[c]{@{}c@{}}\textbf{0.83}\\{[0.72,0.91]}\end{tabular}
& \begin{tabular}[c]{@{}c@{}}5.01\\{[3.93,6.80]}\end{tabular}
& \begin{tabular}[c]{@{}c@{}}0.13\\{[0.12,0.14]}\end{tabular}
& \begin{tabular}[c]{@{}c@{}}\textbf{0.95}\\{[0.92,0.98]}\end{tabular}
& \begin{tabular}[c]{@{}c@{}}\textbf{0.81}\\{[0.71,0.90]}\end{tabular}
& \begin{tabular}[c]{@{}c@{}}\textbf{4.90}\\{[3.85,6.64]}\end{tabular}
& \begin{tabular}[c]{@{}c@{}}0.13\\{[0.12,0.14]}\end{tabular}
\\

& \gls{mcde}
& 0.12
& \begin{tabular}[c]{@{}c@{}}\textbf{0.96}\\{[0.92,0.98]}\end{tabular}
& \begin{tabular}[c]{@{}c@{}}0.79\\{[0.65,0.90]}\end{tabular}
& \begin{tabular}[c]{@{}c@{}}\textbf{6.43}\\{[4.83,9.28]}\end{tabular}
& \begin{tabular}[c]{@{}c@{}}\textbf{0.12}\\{[0.12,0.13]}\end{tabular}
& \begin{tabular}[c]{@{}c@{}}0.83\\{[0.76,0.89]}\end{tabular}
& \begin{tabular}[c]{@{}c@{}}0.40\\{[0.25,0.59]}\end{tabular}
& \begin{tabular}[c]{@{}c@{}}3.25\\{[2.26,4.96]}\end{tabular}
& \begin{tabular}[c]{@{}c@{}}0.13\\{[0.12,0.14]}\end{tabular}
\\
\hline

\end{tabular}
}
\endgroup
\end{table*}

\end{document}